%% file: main.tex
\documentclass{article} 
\usepackage{iclr2027_conference,times}

\input{math_commands.tex}

\usepackage{hyperref}
\usepackage{url}
\usepackage{graphicx}
\usepackage{booktabs}
\usepackage{multirow}
\usepackage{enumitem}
\usepackage{titletoc}

\title{
\parbox{\textwidth}{
\centering
Overcoming Data Imbalances with Expert Selection Biases
}
}

\author{
\parbox{\textwidth}{
\centering
Ayush K. Tarun\thanks{Corresponding author.}
\quad
Clara Meister
\quad
Antoine Bosselut \\
EPFL \\
\texttt{\{ayush.tarun,clara.meister,antoine.bosselut\}@epfl.ch}
}
}

\iclrfinalcopy 
\begin{document}

\maketitle

\fancyhead{}
\lhead{Preprint}

\input{sections/abstract}
\input{sections/introduction}
\input{sections/related_work}
\input{sections/method}
\input{sections/experiments}

\input{sections/results}
\input{sections/ablations}
\input{sections/conclusion}
\input{sections/acknowledgements}

\bibliography{iclr2027_conference}
\bibliographystyle{iclr2027_conference}

\input{sections/appendix}

\end{document}

%% file: math_commands.tex
\usepackage{amsmath,amsfonts,bm}

\def\eqref#1{equation~\ref{#1}}

\def\1{\bm{1}}

\def\ve{{\bm{e}}}

\def\vh{{\bm{h}}}

\def\vs{{\bm{s}}}

\def\vu{{\bm{u}}}

\DeclareMathAlphabet{\mathsfit}{\encodingdefault}{\sfdefault}{m}{sl}
\SetMathAlphabet{\mathsfit}{bold}{\encodingdefault}{\sfdefault}{bx}{n}

\def\sA{{\mathbb{A}}}

\def\sC{{\mathbb{C}}}

\def\sP{{\mathbb{P}}}

\newcommand{\R}{\mathbb{R}}



%% file: sections/abstract.tex
\begin{abstract}
    Mixture-of-Experts (MoE) models have become the standard architecture for frontier LLMs. The multiple parallel experts in each transformer block theoretically enable expert specialization according to categories such as domains, languages, or modalities during pretraining. However, when data is imbalanced across categories, minority categories can become concentrated in a small fraction of the expert pool or spread inefficiently across them, effectively limiting their available capacity. In this work, we demonstrate that category-conditioned expert-selection biases can preferentially route tokens from different categories toward specialized expert pools. By applying biases based on input category, we provide a simple way to induce soft category-aligned specialization. This intervention encourages categories toward distinct expert subsets while retaining cross-category sharing when useful. More importantly, it enables the over-allocation of expert capacity to categories that would otherwise be disadvantaged by data scarcity. We observe performance improvements across modalities, low-resource languages, and code in models up to 10.6B parameters. At our smallest and largest model scales, these gains are equivalent to allocating $18.5\%$ and $2.7\%$ additional baseline training tokens, respectively. Overall, our results show that expert biases can serve as a lightweight control for category-level expert allocation, making room for low-resource categories while retaining standard MoE inference.
\end{abstract}

%% file: sections/introduction.tex
\section{Introduction}
Mixture-of-Experts (MoE) models offer the potential for \textit{specialization}, where different experts are used to process different data categories, e.g., languages, modalities, or domains. 
Such specialization would increase effective modeling capacity by reducing training interference between heterogeneous data categories when there are no shared representations to be learned. 
However, category-aligned specialization does not reliably emerge under standard training \citep{jiang2024mixtral,wang2026mythexpertspecializationmoes}, and capacity may be asymmetrically allocated due to data imbalance even when it does. For example, in multimodal MoEs, \citet{mustafa2022multimodal} show that tokens from a minority modality can collapse onto a single expert and saturate its capacity. Recent analyses of multilingual MoEs similarly report concentrated routing for underrepresented languages \citep{joseph2026mixture}.

Prior work has explored methods to explicitly encourage or enforce category-aware specialization through dedicated experts \citep{gururangan-etal-2022-demix}, category-guided routing \citep{zhao2024sparse}, and objectives that explicitly encourage category-level routing distributions to diverge \citep{ICLR2026_0c8f0ede}. These approaches establish that category information can meaningfully shape expert specialization during pretraining, and suggest that expert specialization could provide a mechanism for separating heterogeneous data categories, potentially reducing the competition for modeling capacity between them. 
However, these results do not establish that category-aware specialization can be used to aid model performance for data-scarce categories where this separation would be most consequential. For these low-resource categories, training alongside higher-resource categories degrades performance relative to training on the category alone, as reported for multimodal \citep{aghajanyan2023scaling, huang2022modality} and multilingual \citep{longpre2025atlas} settings.

\begin{figure}[t]
    \centering
    \includegraphics[width=\textwidth]{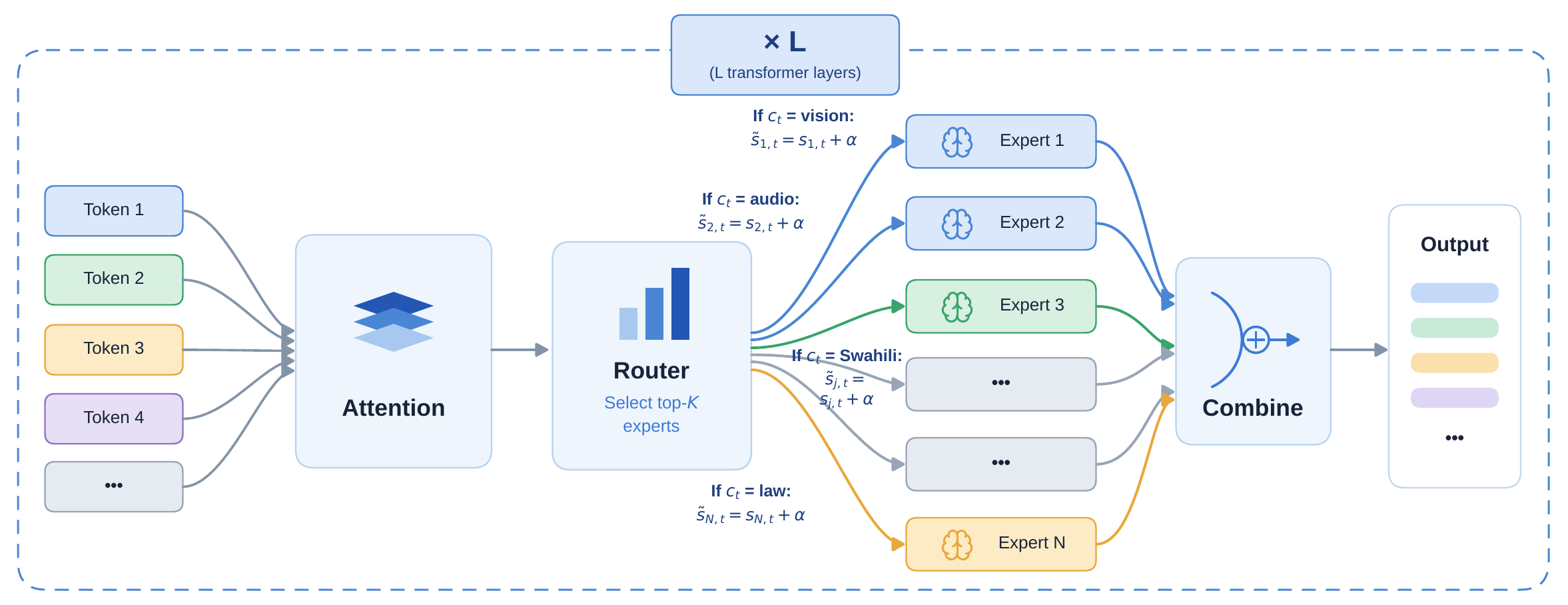}
    \caption{Given an input category, \textbf{COAX} adds a category-conditioned bias $\alpha$ to the routing scores of a designated subset of experts before top-$k$ selection. This bias softly encourages tokens from the same category to route to a common expert set, promoting category-specific specialization, while leaving all experts available when the learned routing scores sufficiently favor them. The mechanism is applied across MoE layers during pretraining, but the biases can be removed at inference time.}
    \label{fig:expert-biasing}
\end{figure}

In this work, we investigate whether category-aligned expert selection alone can benefit low-resource categories. We propose \textbf{COAX} (\textbf{C}ategory-\textbf{O}riented \textbf{A}llocation of e\textbf{X}perts; Figure~\ref{fig:expert-biasing}), which repurposes the per-expert router bias that \citet{wang2024auxiliary} introduced for load balancing.
During pretraining, COAX adds a positive bias to the router probabilities of a designated subset of experts for each category. Rather than biasing experts according to their underuse, COAX biases based on data category, allowing a model developer to explicitly control the fraction of experts designated for a category, what we term the \textit{preferred expert share}. This fraction can be set higher than the category's fraction of the training tokens, e.g., a language that makes up $1\%$ of the training tokens can be assigned 4 of 128 routed experts ($3.1\%$), allowing COAX to overcome data imbalances. 
Crucially, COAX represents a \textit{soft} bias. All experts remain accessible to the router, allowing sufficiently strong learned routing scores to override the category bias, and preserving the possibility of cross-category knowledge sharing. 
The bias only needs to be applied during pretraining. 
At inference, the bias can be removed, simplifying the model to a standard top-$K$ MoE, and preempting the need to slow inference by inducing category labels for input data. 

Across modalities, languages, and code, we find that this simple intervention consistently improves low-resource performance, often with little to no degradation (and sometimes improvements) for higher-resource categories. 
We summarize our contributions as follows:

\begin{itemize}[leftmargin=1.2em, itemsep=1pt, topsep=2pt]
    \item We introduce COAX, which uses expert biases as category-conditioned routing priors. By biasing expert share per category, COAX enables \textit{preferred expert share}, where the model is encouraged to give low-resource categories disproportionately large expert subsets without increasing per-token active compute.
    
    \item We evaluate COAX across modalities, languages, and domains at three model scales up to 10.6B parameters. In our 345M-A147M and 10.6B-A1.2B models, the improvements are equivalent to $18.5\%$ and $2.7\%$ additional training tokens, respectively. The largest gains are concentrated in low-resource categories for multilinguality and multimodality, and we further show that COAX induces functionally localized expert specialization. COAX also provides additional gains when combined with conventional data upsampling.
    
    \item COAX changes only the pretraining dynamics and requires no modifications to the standard MoE inference stack. No explicit category bias is needed at inference, meaning there is no additional inference-time overhead and no requirement for category labels. Its training overhead decreases sharply with scale, reaching $\leq1.3\%$ at 10.6B-A1.2B.
\end{itemize}

%% file: sections/related_work.tex
\section{Related Work}

\textbf{MoE routing, load balancing, and specialization.} Sparse MoEs require mechanisms that prevent expert under-utilization and routing collapse \citep{shazeer2017outrageously}. More recently, expert-wise routing biases have been used for load balancing by dynamically adjusting each expert's selection bias according to its load \citep{wang2024auxiliary,deepseekai2025deepseekv3technicalreport}. Separately, recent analyses question whether standard MoEs naturally develop human-interpretable expertise, suggesting that routing largely reflects representation geometry rather than semantic categories \citep{wang2026mythexpertspecializationmoes}. Memory-Aware Routing tries to tackle this semantic incongruence by encouraging consistent token-expert assignments using experts' historical preferences \citep{hou-etal-2026-pseudo}. 

\textbf{Category-aligned expert specialization.}
DEMix assigns domains to dedicated experts \citep{gururangan-etal-2022-demix}, while Lingual-SMoE introduces hierarchical language-guided routing with language-specific expert groups \citep{zhao2024sparse}. In multimodal settings, LIMoE finds modality-specific routing strategies to perform comparably to joint routing \citep{mustafa2022multimodal}. SMoES encourages soft modality specialization through layer-wise modality scores, adaptive expert binning, and mutual-information regularization in a LLaVA-style adaptation setup initialized from pretrained MoE backbones \citep{bo2026smoes}. Most closely, EDL \citep{ICLR2026_0c8f0ede} uses domain labels to encourage specialization by maximizing the divergence between domain-level routing distributions. COAX instead directly specifies a soft preferred expert subset and exposes the size of this subset as an explicit allocation control. 

\textbf{Low-resource and long-tailed MoEs.} Prior work improves low-resource MoE performance by increasing expert sharing between low- and high-resource languages, either through inference-time routing interventions \citep{bandarkar2026multilingual,chen2026understanding} or routing-alignment objectives \citep{dong-etal-2026-sara}. Other approaches allocate language-specific parameters after pretraining \citep{li-yanaka-2026-neuronmoe}. Closest to our allocation perspective, LTDR increases the number of experts activated for rare visual tokens \citep{cai2025long}. COAX instead keeps the number of active experts per token fixed and changes the size of the expert subset towards which a category is preferentially biased during pretraining, allowing low-resource categories to be encouraged toward a larger fraction of experts than their prevalence in the data. 

%% file: sections/method.tex
\section{Method}
\label{sec:method}

\subsection{Preliminaries: Sparse MoE Routing}
\label{sec:preliminaries}

Consider a Mixture-of-Experts (MoE) layer with $N$ routed experts, of which $K$ are activated for each token. We omit the layer index for clarity, as the same routing operation is applied independently at each MoE layer. Given the input representation $\vu_t \in \R^d$ of token $t$ to this MoE layer, the router computes its affinity with expert $i$ as
\begin{equation}
    s_{i,t}
    =
    G\left(\vu_t^\top \ve_i\right),
    \qquad i \in \{1,\ldots,N\},
    \label{eq:router-score}
\end{equation}
where $\ve_i \in \R^d$ is the learned router representation of expert $i$ and $G(\cdot)$ denotes the nonlinear gating function. Let $\vs_t=(s_{1,t},\ldots,s_{N,t})$ collect the routing scores for token $t$. Standard top-$K$ routing selects the $K$ experts with the highest affinities as
\begin{equation}
    \sA_t
    =
    \operatorname{TopK}(\vs_t,K),
    \label{eq:standard-topk}
\end{equation}
and weights each selected expert by its normalized affinity,
\begin{equation}
    g_{i,t}
    =
    \frac{
        s_{i,t}\,\1\{i\in\sA_t\}
    }{
        \sum_{j\in\sA_t}s_{j,t}
    }.
    \label{eq:standard-gating}
\end{equation}
The output of the routed MoE block is then
\begin{equation}
    \vh_t
    =
    \vu_t
    +
    \sum_{i=1}^{N}
    g_{i,t}\operatorname{FFN}_i(\vu_t).
    \label{eq:moe-output}
\end{equation}
COAX modifies only the routing among routed experts, and therefore, shared experts \citep{dai2024deepseekmoe} operate unchanged.

\subsection{COAX: Category-Oriented Allocation of Experts}
\label{sec:COAX}

Expert-bias-based load balancing \citep{wang2024auxiliary} introduces an expert-wise bias before top-$K$ selection and dynamically adjusts it according to expert utilization. We repurpose this mechanism, determining these biases based on the input category rather than expert underuse. Figure~\ref{fig:expert-biasing} provides a visual overview of COAX.

Let $c_t \in \sC$ denote the category associated with token $t$. For each category $c$, we associate a designated subset of routed experts
\begin{equation}
    \sP_c \subseteq \{1,\ldots,N\},
    \qquad
    M_c = |\sP_c|.
    \label{eq:preferred-experts}
\end{equation}

where $M_c$ is the size of the preferred expert pool for a particular category $c$. For a fixed bias $\alpha \geq 0$, shared across categories, we define the category-conditioned expert bias
\begin{equation}
    b_{c,i}
    =
    \alpha\,\1\{i\in\sP_c\}.
    \label{eq:COAX-bias}
\end{equation}
During pretraining, the scores used for expert selection become
\begin{equation}
    \tilde{s}_{i,t}
    =
    s_{i,t} + b_{c_t,i},
    \label{eq:COAX-score}
\end{equation}
yielding the selected expert set
\begin{equation}
    \tilde{\sA}_t
    =
    \operatorname{TopK}(\tilde{\vs}_t,K),
    \qquad
    \tilde{\vs}_t
    =
    (\tilde{s}_{1,t},\ldots,\tilde{s}_{N,t}).
    \label{eq:COAX-topk}
\end{equation}

Following \citet{wang2024auxiliary}, the category bias affects only expert selection. Once the top-$K$ experts have been selected, their contributions are weighted using the original, unbiased routing scores:
\begin{equation}
    g^{\textsc{COAX}}_{i,t}
    =
    \frac{
        s_{i,t}\,\1\{i\in\tilde{\sA}_t\}
    }{
        \sum_{j\in\tilde{\sA}_t}s_{j,t}
    }.
    \label{eq:COAX-gating}
\end{equation}
COAX therefore acts as a \emph{soft} routing prior rather than a hard partition. Tokens from category $c$ are encouraged toward $\sP_c$, but every expert remains accessible. An expert outside $\sP_c$ can still be selected when its learned routing score sufficiently outweighs the added bias. Likewise, experts in $\sP_c$ remain available to tokens from other categories. This preserves cross-category knowledge sharing while encouraging category-aligned specialization. Since COAX modifies only expert selection during pretraining, the category-conditioned bias can be removed at inference. The resulting model uses standard top-$K$ routing and requires neither category labels nor changes to the inference procedure.

\paragraph{Preferred expert share.}
Beyond specialization, COAX lets us explicitly specify a category's preferred share of the expert pool independently of its share of the training data. We define $q_c$ as the \emph{preferred expert share} of $c$ (i.e., the fraction of routed experts toward which category $c$ is biased):
\begin{equation}
    q_c = \frac{M_c}{N}
    \label{eq:expert-fraction}
\end{equation}
We can also define the \emph{relative expert share} of category $c$: the preferred expert share of $c$ divided by its token share $p_c$ (i.e., the fraction of the training mixture belonging to category $c$):
\begin{equation}
    r_c
    =
    \frac{q_c}{p_c}
    =
    \frac{M_c}{N p_c}.
    \label{eq:parametric-over-allocation}
\end{equation}
Here, $r_c>1$ indicates that category $c$ is biased toward a larger fraction of the expert pool than its fraction of the training data tokens. For example, a category comprising $2.5\%$ of the training mixture can be biased toward $10\%$ of the routed experts, corresponding to $r_c=4$.

Note that increasing $M_c$ does not change the number $K$ of experts activated for an individual token, meaning per-token active compute stays the same even with increased expert capacity. 

%% file: sections/experiments.tex
\section{Experimental Setup}
\label{sec:experimental-setup}

We evaluate COAX under three common settings where pretraining data features low-resource categories: multimodal, multilingual and multi-domain.
For each setting, we train MoE models at three sizes: 345M-A147M, 3B-A654M, and 10.6B-A1.2B.

\label{sec:models} \paragraph{Models.} All models consist of 32 layers and use Grouped Query Attention (GQA) \citep{ainslie2023gqa}, one shared expert, top-$K=8$ routing, and tied embeddings. We use SwiGLU activations and QK-Norm for training stability \citep{chameleonteam2025chameleonmixedmodalearlyfusionfoundation}. The router uses a sigmoid activation. The remaining architecture parameters vary with model scale and are summarized in Appendix ~\ref{app:model-config}.

\label{sec:tokenizers} \paragraph{Tokenizers.} For text, we use the tokenizer from Mistral-Nemo \citep{mistralnemo} (vocabulary size $131$K), Emu3.5 tokenizer \citep{cui2025emu35nativemultimodalmodels} for images (vocabulary size $131$K)  and WavTokenizer (vocabulary size $4$K) \citep{ji2024wavtokenizer} with a token rate of $40$ tokens/second for audio.

\label{sec:data-evaluation} \paragraph{Datasets and Evaluation.} 
We train and evaluate on held-out splits from high-quality text, multilingual, vision, and audio datasets. Refer to Appendix~\ref{app:dataset} for full details.


\label{sec:training} \paragraph{Training.} All models are trained on 64 NVIDIA GH200 GPUs with 96\,GB of VRAM using the Megatron-LM \citep{shoeybi2019megatron} framework. All models are trained on 10B tokens unless otherwise stated. We set the learning rate according to \citet{li2025predictablescaleistep} and a global batch size of 256. We use the WSD scheduler \citep{wen2024understanding}, warmup for 3\% of the steps, and linearly decay the LR to 0 in the last 10\% steps. All experiments use a sequence length of 4096. The 345M-A147M and 3B-A654M models use a micro-batch size of 4, while the 10.6B-A1.2B model uses a micro-batch size of 2. The two larger models additionally use expert parallelism with degree 2. We also use auxiliary load-balancing loss with coefficient $1e^{-3}$. 
Unless otherwise stated, COAX $\alpha=0.03$ (Eq. \ref{eq:COAX-bias}) for the biased runs, and expert subsets ($\sP_c$, Section \ref{sec:COAX}) are mutually exclusive across categories.


%% file: sections/results.tex
\section{Results}
\label{sec:results}

\begin{figure}[t]
    \centering
    \includegraphics[width=\textwidth]{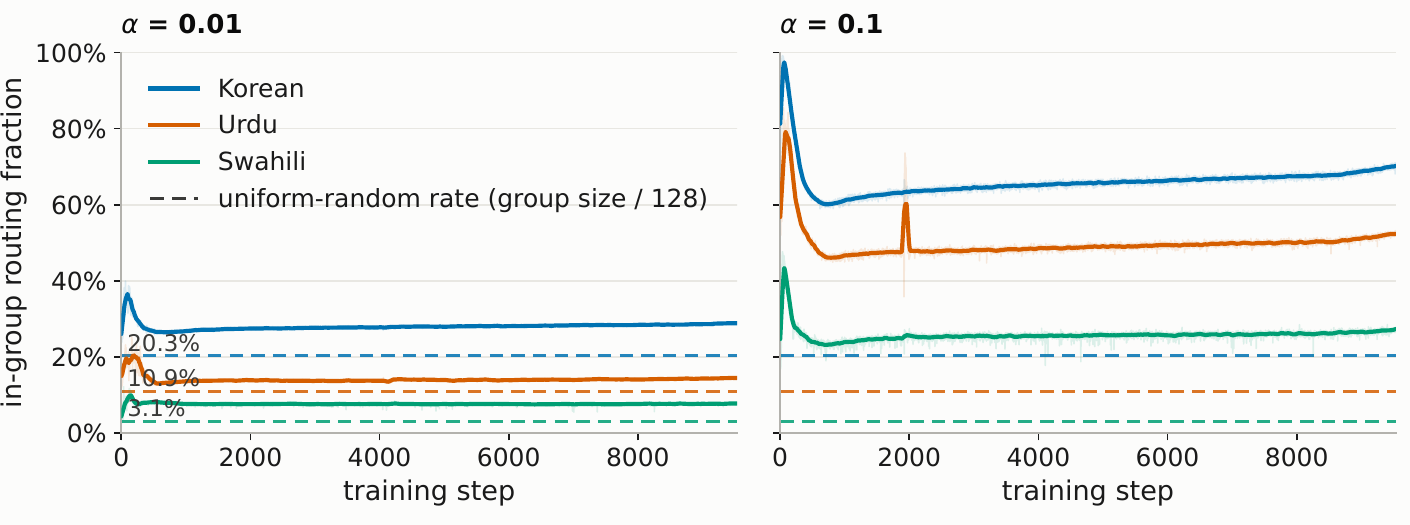}
    \caption{\textbf{COAX effectively \textit{coaxes} category-aligned routing.} In-group routing fraction: the fraction of a token's top-$K$ selected experts that belong to its designated subset $\mathcal{P}_{c_t}$. Dashed lines show the uniform-random routing rate $q_c$ which would be the expected fraction for an unbiased baseline. Larger bias strengths ($\alpha$) consistently produce stronger routing toward the designated expert subset.}
    \label{fig:in-group-routing-fraction}
\end{figure}
\paragraph{COAX effectively \textit{coaxes} category-aligned routing.}
Figure~\ref{fig:in-group-routing-fraction} shows that COAX {nudges} tokens toward their designated expert subsets $\sP_{c_t}$. Increasing $\alpha$ leads to substantially stronger in-group routing, providing direct control over the degree of category-aligned specialization. The in-group fractions peak early in training, when the learned routing scores are still close to uniform and the additive bias therefore has a comparatively large influence on expert selection. They subsequently decrease before gradually increasing again over training.

We next evaluate whether this induced specialization translates into improved performance across multilingual, multimodal and domain mixtures. 

\subsection{Multilinguality}
\label{sec:results-multilingual}

Our multilingual mixture consists of 86.42\% English, 10\% Korean, 2.73\% Urdu, and 0.85\% Swahili. We use five seeds for the 345M-A147M model and four seeds for the remaining scales. We use a relative expert share of 2, 4, and 4 for Korean, Urdu, and Swahili, respectively. English receives the residual expert subset, $r_c\approx0.8$.

\begin{table}[t]
  \centering\small
  \setlength{\tabcolsep}{6pt}
  \begin{tabular}{lccccc}
    \toprule
    \textbf{Model} & \textbf{Inf. Bias} & \textbf{English} & \textbf{Korean} & \textbf{Urdu} & \textbf{Swahili} \\
    \midrule
    \multirow{2}{*}{345M-A147M} & No & $+0.05$ & $\mathbf{-0.01}$ & $\mathbf{-0.23}$ & $\mathbf{-0.55}^{**}$ \\
     & Yes & $\mathbf{-0.02}$ & $\mathbf{-0.08}$ & $\mathbf{-0.28}$ & $\mathbf{-0.59}^{**}$ \\
    \midrule
    \multirow{2}{*}{3B-A654M} & No & $\mathbf{-0.01}$ & $\mathbf{-0.11}$ & $\mathbf{-0.31}^{*}$ & $\mathbf{-0.40}^{**}$ \\
     & Yes & $\mathbf{-0.09}^{*}$ & $\mathbf{-0.18}$ & $\mathbf{-0.38}^{*}$ & $\mathbf{-0.44}^{**}$ \\
    \midrule
    \multirow{2}{*}{10.6B-A1.2B} & No & $+0.02$ & $\mathbf{-0.00}$ & $\mathbf{-0.10}^{*}$ & $\mathbf{-0.30}^{**}$ \\
     & Yes & $\mathbf{-0.09}$ & $\mathbf{-0.13}^{*}$ & $\mathbf{-0.22}^{**}$ & $\mathbf{-0.41}^{**}$ \\
    \bottomrule
  \end{tabular}
  \caption{\textbf{Mean relative change in validation loss ($\Delta\%$) against unbiased baselines trained with matched seeds (negative is better).} \emph{'Inf. Bias'} denotes whether the category-conditioned bias is retained during inference. \textbf{Bold} indicates improvements over the unbiased baseline ($\Delta<0$). $^{*}$ and $^{**}$ denote paired differences whose $90\%$ and $95\%$ confidence intervals respectively exclude zero. Our results demonstrate that COAX more significantly helps lower-resource languages with minimal impact to higher-resource languages.}
  \label{tab:ml-results}
\end{table}


\paragraph{Benefits increase as languages become lower-resource.} Table~\ref{tab:ml-results} shows that COAX improves low-resource-language performance in nearly every setting. The gains become progressively larger as a language's share of the training data decreases. Korean, which constitutes $10\%$ of the training mixture, receives relatively modest improvements. Urdu, meanwhile, at $2.73\%$, improves more consistently, and Swahili, at only $0.85\%$, improves substantially and significantly at every model scale. This trend supports the central motivation of COAX: \textbf{higher relative expert share can \textit{make room} for categories that occupy only a small fraction of the training mixture.} Beyond these loss improvements, COAX also induces functionally localized experts: disabling the four Swahili-designated experts increases Swahili loss by $107.1\%$, compared with only $9.3\%$ when disabling the four experts most frequently used by Swahili in the unbiased baseline (Appendix~\ref{app:modularity}). 

Retaining the bias during inference generally provides additional gains, but removing it preserves most of the benefit. Thus, the category information used by COAX can remain a training-time signal without requiring category-conditioned inference. Moreover, Appendix~\ref{app:rc-ablation} shows that performance is relatively insensitive to the precise expert share once $r_c\geq2$ for Swahili, suggesting that the benefits do not require finely tuning $r_c$. 

\label{sec:8_language_result}\paragraph{COAX extends to a more practical multilingual mixture.} To test whether these gains persist in a more heterogeneous setting, we additionally train the 10.6B-A1.2B model on a 10B-token mixture of eight languages: English (63.24\%), Chinese (10.00\%), Korean (10.00\%), Khmer (6.00\%), Tamil (4.19\%), Georgian (2.99\%), Urdu (2.73\%), and Swahili (0.85\%). Chinese and Korean use $r_c=1$, Swahili uses $r_c=4$, and the remaining minority languages use $r_c=2$. Table~\ref{tab:ml-practical} shows that COAX improves seven of eight languages while leaving Khmer approximately unchanged. The gains remain substantial for several of the lowest-resource languages: Tamil ($-0.42\%$), Georgian ($-0.58\%$), Urdu ($-0.31\%$), and Swahili ($-0.49\%$), while English also improves slightly ($-0.07\%$). \textbf{Thus, making additional expert capacity available to the long tail need not come at the expense of the dominant language.} Although the relative loss changes are small, Appendix~\ref{app:data-equivalence} shows that their training-equivalent benefit can exceed twice COAX's training overhead. 

Table~\ref{tab:ml-practical} also compares COAX against DEMix-style hard routing (by fixing expert groups) and Expert Divergence Learning (EDL). COAX achieves the best mean relative loss change over the seven minority languages ($-0.286\%$), compared with hard routing ($+0.047\%$) and EDL ($+1.016\%$) highlighting the importance of higher relative expert shares across minority categories along with specialization. Appendices~\ref{app:hard-routing} and~\ref{app:edl} contain comparisons to more variants of these methods.

\begin{table}[t]
    \centering
    \small
    \setlength{\tabcolsep}{5pt}
    \begin{tabular}{lccc|lccc}
        \toprule
        \textbf{Language} & \textbf{COAX} & \textbf{Hard-routing} & \textbf{EDL} &
        \textbf{Language} & \textbf{COAX} & \textbf{Hard-routing} & \textbf{EDL} \\
        \midrule
        English  & $\mathbf{-0.070}$ & $+0.350$ & $+0.047$ &
        Tamil    & $-0.418$ & $\mathbf{-0.423}$ & $+0.400$ \\
        Chinese  & $\mathbf{-0.103}$ & $+1.994$ & $+0.222$ &
        Georgian & $\mathbf{-0.576}$ & $+0.228$ & $+1.257$ \\
        Korean   & $\mathbf{-0.104}$ & $+1.679$ & $+0.091$ &
        Urdu     & $\mathbf{-0.308}$ & $-0.101$ & $+0.926$ \\
        Khmer    & $-0.003$ & $\mathbf{-4.090}$ & $+1.193$ &
        Swahili  & $\mathbf{-0.486}$ & $+1.039$ & $+3.020$ \\
        \bottomrule
    \end{tabular}
    \caption{
    \textbf{Relative change in validation loss ($\Delta\%$) for the 10.6B-A1.2B model trained on eight languages against an unbiased baseline  (negative is better).} Columns correspond to our proposed method COAX, DEMix-style hard routing (Hard-routing) and Expert Divergence Learning (EDL). \textbf{Bold} denotes the best result for each language. Hard-routing and EDL induce category-aligned specialization without consistent performance benefits in this setting, highlighting the importance of COAX's ability to allocate different preferred expert shares across categories.}
    \label{tab:ml-practical}
\end{table}


We further stress-test these multilingual results:

\begin{itemize}[leftmargin=1.2em, itemsep=1pt, topsep=2pt]
    \item \textbf{The gains persist with longer training.}
    Although the 10.6B-A1.2B models are trained on only $\approx 8$ tokens per active parameter, Appendix~\ref{app:training-trajectory} shows that COAX's advantage remains stable through 100B tokens, arguing against an under-training artifact.

    \item \textbf{Reducing majority data does not reproduce COAX.}
    Simply reducing the amount of English training data does not recover the same gains for low-resource languages (Appendix~\ref{app:en-data-cut}).

    \item \textbf{COAX complements data upsampling.}
    COAX provides additional gains when combined with conventional upsampling of low-resource language data (Appendix~\ref{app:additivity}).

\end{itemize}

\subsection{Multimodality}
\label{sec:results-multimodal}

\begin{table}[t]
  \centering\small
  \setlength{\tabcolsep}{5pt}
  \begin{tabular}{lccccccc}
    \toprule
    \textbf{Model} & \textbf{Inf. Bias} & \textbf{T} & \textbf{I2T} & \textbf{T2I} & \textbf{AA} & \textbf{AT} & \textbf{ATA} \\
    \midrule
     \multirow{2}{*}{345M-A147M} & No & $\mathbf{-0.78}^{*}$ & $\mathbf{-0.72}^{**}$ & $\mathbf{-0.56}^{**}$ & $\mathbf{-0.48}$ & $\mathbf{-0.38}$ & $\mathbf{-0.65}$ \\
     & Yes & $\mathbf{-0.88}^{*}$ & $\mathbf{-0.76}^{**}$ & $\mathbf{-0.59}^{**}$ & $\mathbf{-0.50}^{*}$ & $\mathbf{-0.39}$ & $\mathbf{-0.67}$ \\
    \midrule
     \multirow{2}{*}{3B-A654M} & No & $+0.23^{**}$ & $\mathbf{-0.25}^{**}$ & $\mathbf{-0.19}^{**}$ & $+0.04$ & $+0.11$ & $+0.09$ \\
     & Yes & $+0.18^{**}$ & $\mathbf{-0.29}^{**}$ & $\mathbf{-0.22}^{**}$ & $+0.02$ & $+0.09$ & $+0.08$ \\
    \midrule
     \multirow{2}{*}{10.6B-A1.2B} & No & $\mathbf{-0.05}$ & $\mathbf{-0.25}^{*}$ & $\mathbf{-0.18}^{*}$ & $\mathbf{-0.08}$ & $\mathbf{-0.09}$ & $\mathbf{-0.09}$ \\
     & Yes & $\mathbf{-0.13}$ & $\mathbf{-0.31}^{**}$ & $\mathbf{-0.23}^{**}$ & $\mathbf{-0.12}^{*}$ & $\mathbf{-0.12}^{*}$ & $\mathbf{-0.13}$ \\
    \bottomrule
  \end{tabular}
  \caption{\textbf{Mean relative change in validation loss ($\Delta\%$) against unbiased baselines trained with matched seeds (negative is better).} Columns correspond to held-out text (T), image-to-text captioning (I2T), text-to-image generation (T2I), audio-only (AA), audio-to-text ASR (AT), and interleaved audio-text (ATA) subtasks in validation data. Vision and audio are each biased toward approximately $2\times$ their token share, with text receiving the residual expert subset ($r_c\approx0.7$). \emph{Inf. Bias} denotes whether the category-conditioned bias is retained during inference. \textbf{Bold} indicates improvements over the unbiased baseline ($\Delta<0$). $^{*}$ and $^{**}$ denote paired differences whose $90\%$ and $95\%$ confidence intervals respectively exclude zero. Our results show that COAX improves data-scarce modalities with minimal impact on the higher-resource text modality. 
  }
  \label{tab:mm-results}
\end{table}

We next study COAX's effect on multimodality. Our multimodal training mixture consists of 75\% text, 12.5\% vision, and 12.5\% audio. Text consists entirely of English FineWeb-Edu. Vision is split equally between image-to-text (6.25\%) and text-to-image (6.25\%) sequences. Audio consists of audio-only (6.25\%), ASR (4.84\%), and interleaved audio--text (1.41\%) sequences. We use five seeds for the 345M-A147M model and four seeds for the remaining scales.


\paragraph{COAX improves multimodal performance, most consistently for vision.} Table~\ref{tab:mm-results} shows that with $r_c=2$, COAX improves all validation sets at the smallest and largest model scales. The gains are largest (and many are significant) for the smallest 345M-A147M model. At 3B-A654M, COAX continues to improve both vision objectives, although these gains come with small regressions on text and audio. Across all three scales, vision exhibits the most consistent improvements, with significant reductions in both I2T and T2I loss. Notably, at the 345M-A147M and 10.6B-A1.2B scales, text also improves despite receiving a smaller fraction of the expert pool. A possible explanation is that biasing vision and audio toward larger preferred expert pools reduces their interference with the experts that text data primarily relies on. In Table~\ref{tab:downstream-mm} of Appendix~\ref{app:downstream}, we show these validation-loss benefits are also reflected in ASR and captioning accuracy.

\paragraph{Most gains persist without inference-time bias.} As with multilinguality, keeping the category bias active during inference generally provides a small additional improvement, which is practical for multimodal inputs, as the token modality is known \textit{a priori}. However, removing the bias at inference time still retains most of COAX's benefits, indicating that the training-time intervention induces useful specialization that persists under standard MoE inference.

\subsection{Domains}
\label{sec:dom-results}


\begin{table}[t]
  \centering\small
  \setlength{\tabcolsep}{6pt}
  \begin{tabular}{llccccc}
    \toprule
    \textbf{Model} & \textbf{Inf. Bias} & \textbf{English} & \textbf{Code} & \textbf{Math} & \textbf{Legal} & \textbf{Medical} \\
    \midrule
    \multirow{2}{*}{345M-A147M} & No & $+0.26$ & $\mathbf{-0.19}$ & $+0.10$ & $+0.04$ & $+0.07$ \\
     & Yes & $+0.14$ & $\mathbf{-0.35}$ & $\mathbf{-0.05}$ & $\mathbf{-0.15}$ & $\mathbf{-0.11}$ \\
    \midrule
    \multirow{2}{*}{3B-A654M} & No & $+0.20^{**}$ & $\mathbf{-0.10}$ & $+0.62$ & $+0.05$ & $+0.09$ \\
     & Yes & $+0.10$ & $\mathbf{-0.25}$ & $\mathbf{-0.12}$ & $\mathbf{-0.09}$ & $\mathbf{-0.06}$ \\
    \midrule
    \multirow{2}{*}{10.6B-A1.2B} & No & $+0.25^{**}$ & $+0.00$ & $+0.10$ & $+0.05$ & $+0.11$ \\
     & Yes & $+0.07$ & $\mathbf{-0.19}^{*}$ & $\mathbf{-0.04}$ & $\mathbf{-0.08}$ & $\mathbf{-0.04}$ \\
    \bottomrule
  \end{tabular}
  \caption{\textbf{Mean relative change in validation loss ($\Delta\%$) against matched-seed unbiased baselines (negative is better)}. English is under-provisioned ($r_c\approx0.7$), code and mathematics use $r_c=2$, and legal and medical use $r_c=4$. \emph{Inf. bias} denotes whether the category-conditioned bias is retained during inference. \textbf{Bold} denotes improvement over the baseline; $^{*}$ and $^{**}$ denote paired differences whose $90\%$ and $95\%$ confidence intervals respectively exclude zero. Our results demonstrate that COAX may help specialized domains when the expert bias is kept at inference time, while removing it only helps code. 
  }
  \label{tab:dom-results}
\end{table}

Our domain mixture contains 85\% general English, 5\% code, 5\% mathematics, and 2.5\% each of legal and medical text. Table~\ref{tab:dom-results} shows a different regime from the multilingual and multimodal settings. Without inference-time bias, code is the only specialist domain that consistently benefits across model scales. When the bias is retained, all four specialist domains improve at every scale, likely because non-code domains (e.g., mathematics, legal, and medical text) share more useful structure with general English.

Finally, Appendix~\ref{app:cpt} studies whether COAX can direct newly added expert capacity toward categories introduced during continual pretraining (CPT). Here, COAX works best for code, followed by vision, while newly introduced languages (Russian and Hindi) slightly underperform vanilla CPT. However, performance on the replay categories improves across all four settings with COAX.

%% file: sections/ablations.tex
\section{Ablations}
\label{sec:ablations}

\begin{table}[b]
    \centering
    \small
    \setlength{\tabcolsep}{7pt}
    \begin{tabular}{lcc|lcc}
        \toprule
        \textbf{Language} & \multicolumn{2}{c|}{$\boldsymbol{\Delta}$\textbf{ loss (\%)}} &
        \textbf{Language} & \multicolumn{2}{c}{$\boldsymbol{\Delta}$\textbf{ loss (\%)}} \\
        \cmidrule(lr){2-3}\cmidrule(lr){5-6}
         & Per-lang. & Eng.\ vs rest &  & Per-lang. & Eng.\ vs rest \\
        \midrule
        English  & $-0.07$ & $\mathbf{-0.15}$ & Tamil    & $-0.42$ & $\mathbf{-0.53}$ \\
        Chinese  & $-0.10$ & $\mathbf{-0.20}$ & Georgian & $-0.58$ & $-0.58$ \\
        Korean   & $-0.10$ & $\mathbf{-0.29}$ & Urdu     & $-0.31$ & $\mathbf{-0.37}$ \\
        Khmer    & $\mathbf{0.00}$  & $+0.19$ & Swahili  & $\mathbf{-0.49}$ & $-0.26$ \\
        \bottomrule
    \end{tabular}
    \caption{\textbf{Fine-grained expert partitioning is not necessary.} Relative change in validation loss for the 10.6B-A1.2B model trained on the eight-language mixture from Section~\ref{sec:8_language_result}; negative is better. \emph{Per-lang.} assigns each language its own expert subset, whereas \emph{Eng.\ vs rest} uses only two groups: English and all non-English languages. Bold denotes the better grouping for each language. The coarser split performs as well as or better than per-language partitioning for six of eight languages, suggesting that much of COAX's benefit comes from separating the dominant language from the long tail rather than separating every minority language from each other.}
    \label{tab:ml-practical-simple}
\end{table}

\label{sec:coarse_grouping}\paragraph{Coarse expert grouping is often sufficient.} Section~\ref{sec:8_language_result} shows that COAX improves performance across a heterogeneous eight-language mixture, but the per-language setup requires choosing a separate relative expert share for each language. To test whether such fine-grained partitioning is necessary, we consider a simpler configuration with only two expert groups: English and all non-English languages. Table~\ref{tab:ml-practical-simple} shows that this coarser grouping performs as well as or better than the per-language setup for six of eight languages, suggesting that COAX does not require meticulous expert partitioning to obtain performance gains. More broadly, the result indicates that much of the benefit may come from reducing interference between the dominant and minority categories, rather than isolating the lower-resource categories from one another. Appendix~\ref{app:no-majority-bias} further supports this by showing that not isolating English erases much of the improvements. 

This simpler configuration is also favorable in training cost. For the 10.6B-A1.2B model, \textbf{it adds only $1.3\%$ training-time overhead for each step (compared to $2.4\%$ with finer expert grouping), while the unbiased baseline requires $2.7\%$ additional training tokens to match its average validation loss}. We further analyze the throughput overhead and training-token equivalence in Appendices~\ref{app:throughput} and~\ref{app:data-equivalence}, respectively.
\label{sec:strength_alpha}\paragraph{Stronger biasing increases specialization, but also train-test mismatch.} We ablate the bias strength $\alpha$ from $0.003$ to $0.3$ in Table~\ref{tab:alpha-ladder}. When the category bias is removed at evaluation time, performance improves up to $\alpha=0.03$, after which the mismatch between biased training and unbiased inference becomes increasingly harmful, becoming catastrophic at $\alpha=0.3$. In contrast, when the bias is retained at evaluation time, minority-language performance improves monotonically with $\alpha$, reaching a mean $\Delta$ loss of $-1.65\%$ at $\alpha=0.3$. However, stronger biasing at inference time comes with increased throughput overhead as load balancing relative to the data distribution becomes more skewed. The slight non-monotonicity at the largest \(\alpha\) values may arise because expert selection begins to saturate. Once the bias largely confines tokens to their category-aligned expert pools, increasing \(\alpha\) more induces only marginal changes to routing and the resulting expert loads.

A simple way to retain stronger specialization without requiring bias at inference is to anneal $\alpha$ to zero. We linearly decay $\alpha=0.3$ during the cooldown phase and recover $76.4\%$ of the minority-language improvement when compared to $\alpha=0.3$ with bias active at inference ($-1.26\%$ vs.\ $-1.65\%$). English still degrades by $+0.57\%$, indicating that annealing substantially reduces the train-test mismatch but does not eliminate the dominant-language trade-off. 
\begin{table}[t]
    \centering
    \small
    \setlength{\tabcolsep}{7pt}
    \begin{tabular}{lccccc}
        \toprule
        & \multicolumn{2}{c}{\textbf{No inference bias} ($\boldsymbol{\Delta}$\textbf{ loss \%})}
        & \multicolumn{2}{c}{\textbf{With inference bias} ($\boldsymbol{\Delta}$\textbf{ loss \%})}
        & \\
        \cmidrule(lr){2-3}\cmidrule(lr){4-5}
        $\boldsymbol{\alpha}$
        & \textbf{English}
        & \textbf{Minority mean}
        & \textbf{English}
        & \textbf{Minority mean}
        & $\boldsymbol{\Delta}$ \textbf{TFLOP/s} \\
        \midrule
        $0.003$ & $-0.02$ & $-0.12$ & $-0.02$ & $-0.12$ & $0.0\%$ \\
        $0.01$  & $-0.10$ & $-0.30$ & $\mathbf{-0.11}$ & $\mathbf{-0.31}$ & $-3.9\%$ \\
        $0.03$  & $-0.03$ & $-0.42$ & $\mathbf{-0.12}$ & $\mathbf{-0.48}$ & $-13.1\%$ \\
        $0.1$   & $+0.38$ & $-0.37$ & $\mathbf{-0.04}$ & $\mathbf{-0.96}$ & $-27.3\%$ \\
        $0.3$   & $+1.88$ & $+19.32$ & $\mathbf{+0.46}$ & $\mathbf{-1.65}$ & $-25.6\%$ \\
        \midrule
        $0.3$, decayed & $+0.57$ & $\mathbf{-1.26}$ & \multicolumn{2}{c}{--} & -- \\
        \bottomrule
    \end{tabular}
    \caption{\textbf{Effect of bias strength.} Relative change in validation loss for English and the mean over the three minority languages: Korean, Urdu, and Swahili, for the 3B-A654M model on the four-language mixture from Section~\ref{sec:results-multilingual}; negative is better. \emph{No inference bias} removes the category bias at inference, whereas \emph{with inference bias} retains the training-time bias. Without validation-time bias, gains peak around $\alpha=0.03$ before train-test mismatch dominates at larger values. Retaining the bias yields increasingly large minority-language gains as $\alpha$ increases, at the cost of greater inference overhead. In the final row, $\alpha=0.3$ is linearly annealed to zero during cooldown, recovering $76.4\%$ of the corresponding minority-language gain without requiring biasing at inference.}
    \label{tab:alpha-ladder}
\end{table}

%% file: sections/conclusion.tex
\section{Conclusion}

We show that expert biases introduced for load balancing can be repurposed to shape category-aligned routing and make a category's preferred expert share an explicit training-time control. Across multimodal and multilingual mixtures, COAX improves low-resource performance while preserving standard MoE inference, and our analyses show that the induced specialization becomes functionally localized in the experts. The method does not require finely tuned expert partitions, can provide additional gains alongside conventional data upsampling, and at the 10.6B scale offers a favorable trade-off between its performance gains and training overhead. Our results therefore suggest that router biases can serve a broader role than balancing expert utilization: they can also be used to make room for the long tail while retaining soft cross-category sharing. Future work could replace manually specified categories with a learned structure and explore systems-aware biasing that better aligns preferred experts with the expert-parallel topology.

%% file: sections/acknowledgements.tex
\section{Acknowledgements}

We thank Madhur Panwar, María Grandury, Gül Altintas, and Deniz Bayazit for their helpful discussions and feedback on our manuscript. We also gratefully acknowledge the support of the Swiss National Science Foundation (No. 215390), the AI2050 program at Schmidt Sciences (Grant \#G-25-69783), Sony Group Corporation, and the Swiss National Supercomputing Center (CSCS) in the form of an infrastructure engineering and development project. This work was also supported as part of the Swiss AI Initiative by a grant from the Swiss National Supercomputing Centre (CSCS) under project IDs a139 and a145 on Alps. Furthermore, this research is with support from Google.org and the Google Cloud Research Credits program for the Gemini Academic Program. This project is also funded by the European Union (ERC, RESPECT-LM, 101222478). Views and opinions expressed are however those of the authors only and do not necessarily reflect those of the European Union or the European Research Council. Neither the European Union nor the granting authority can be held responsible for them. Finally, Ayush acknowledges support from the EPFL AI Center PhD Fellowship Program 2026.

%% file: sections/appendix.tex
\clearpage
\appendix
\onecolumn

\startcontents[appendix]

{\LARGE\bfseries Appendix}\par\vspace{-2mm}

\printcontents[appendix]{}{1}{}
\clearpage

\section{Experimental Details}
\subsection{Model Configurations}
\label{app:model-config}
Table~\ref{tab:model-configs} summarizes the architecture parameters for the three model scales used in our experiments. We use highly sparse MoE configurations, with many routed experts but only a small subset activated per token, reflecting the design of contemporary MoE models \citep{qwen3_8, kimiteam2026kimik3openfrontier}.

\begin{table}[b]
    \centering
    \small
    \begin{tabular}{lccccc}
        \toprule
        \textbf{Model} & \textbf{Hidden dim.} & \textbf{Expert dim.} &
        \textbf{\# Experts} & \textbf{\# Heads} & \textbf{\# KV Heads} \\
        \midrule
        345M-A147M  & 384  & 96  & 64  & 16 & 4 \\
        3B-A654M    & 1280 & 160 & 128 & 20 & 5 \\
        10.6B-A1.2B & 2048 & 192 & 256 & 16 & 4 \\
        \bottomrule
    \end{tabular}
    \caption{Architecture configurations of the MoE models used in our experiments. All models use 32 layers, one shared expert, and activate $K=8$ routed experts per token.}
    \label{tab:model-configs}
\end{table}

\subsection{Datasets and Evaluation}
\label{app:dataset}
We prioritize high-quality datasets to obtain reliable training and evaluation signals at our experimental scales. English text is sampled from FineWeb-Edu \citep{lozhkov2024fineweb-edu}, while multilingual data is drawn from a quality-filtered subset of FineWeb-2 \citep{penedo2025fineweb2pipelinescale}. For vision, we use the LLaVA-OneVision mid-training dataset \citep{an2025llavaonevision15fullyopenframework}, with an equal mixture of image-caption and reversed caption-image sequences (to simulate image-generation). For audio, we sample audio-only data from Unsupervised People's Speech (UPS; \citealp{USP}), ASR data from GigaSpeech \citep{chen21o_interspeech}, and interleaved audio-text data from the 'large' subset of Libriheavy \citep{kang2024libriheavy}. For our domain experiments, we consider four domains alongside general English text: code from StarCoder \citep{li2023starcoder}, mathematics from the FineMath-3+ subset \citep{allal2025smollm2smolgoesbig}, legal text from the English portion of Legal-mC4 \citep{niklaus-etal-2024-multilegalpile}, and biomedical text from Biomed-Enriched \citep{touchent-etal-2026-biomed}. Evaluation is done on 100M held-out tokens from each data category except Swahili and audio-text interleaved, which have 20M token held-out sets.

\section{Additional Evaluations}
We complement the main experiments along two axes: task-level evaluations beyond validation loss, and a continual-pretraining setting in which COAX is introduced after initial pretraining.

\subsection{Task-Level Evaluation}
\label{app:downstream}

\paragraph{Multilingual downstream benchmarks are underpowered at our training scale.} Our primary multilingual evaluation uses held-out loss because the per-language token budgets are too small for standard downstream benchmarks to reliably resolve the observed differences. Table~\ref{tab:downstream} reports MultiBLiMP on the 10.6B-A1.2B eight-language setup (Section~\ref{sec:8_language_result}) for the three available low-resource languages. COAX-baseline differences are small and inconsistent across unnormalised and length-normalised scoring, while both models retain high absolute accuracy. Belebele is similarly uninformative, with baseline accuracies of $21.6$--$27.0\%$ around the $25\%$ chance level, while SIB-200 produces an identical $25.10\%$ accuracy for both models.

\begin{table}[t]
  \centering\small
  \setlength{\tabcolsep}{6pt}
  \begin{tabular}{lrccrccr}
    \toprule
    & & \multicolumn{3}{c}{\textbf{Unnormalised}} & \multicolumn{3}{c}{\textbf{Length-normalised}} \\
    \cmidrule(lr){3-5}\cmidrule(lr){6-8}
    \textbf{Language} & \textbf{$n$} & \textbf{Base} & \textbf{COAX} & \textbf{$\Delta$} & \textbf{Base} & \textbf{COAX} & \textbf{$\Delta$} \\
    \midrule
    Tamil    & $382$ & $98.17$ & $97.91$ & $-0.26$ & $90.05$ & $91.62$ & $+1.57$ \\
    Georgian & $204$ & $96.57$ & $96.08$ & $-0.49$ & $91.67$ & $92.16$ & $+0.49$ \\
    Urdu     & $550$ & $96.55$ & $96.91$ & $+0.36$ & $95.45$ & $95.09$ & $-0.36$ \\
    \bottomrule
  \end{tabular}
  \caption{\textbf{MultiBLiMP shows no consistent COAX effect at our training scale.} Accuracy (\%) for the 10.6B-A1.2B eight-language models evaluated without inference-time bias; higher is better. Differences are small and inconsistent across the two scoring schemes, while both models retain high absolute performance.}
  \label{tab:downstream}
\end{table}

\paragraph{Multimodal improvements transfer to task-level metrics.} We condition on the input and measure top-$k$ accuracy only over the reference transcript or caption from our held-out set. Table~\ref{tab:downstream-mm} shows that COAX improves all twelve token- and word-level metrics for a 345M-A147M model, with absolute gains of $0.66$-$1.88$ points. For example, ASR top-5 word accuracy increases from $47.50\%$ to $49.38\%$, while captioning top-5 word accuracy increases from $78.52\%$ to $79.57\%$. Thus, the validation-loss improvements in Table~\ref{tab:mm-results} are also reflected in task-aligned predictions rather than being confined to the loss metric.

\begin{table}[b]
  \centering\small
  \setlength{\tabcolsep}{6pt}
  \begin{tabular}{llcccccc}
    \toprule
    & & \multicolumn{3}{c}{\textbf{Token-level}} & \multicolumn{3}{c}{\textbf{Word-level}} \\
    \cmidrule(lr){3-5}\cmidrule(lr){6-8}
    \textbf{Task} & & \textbf{top-1} & \textbf{top-5} & \textbf{top-10} & \textbf{top-1} & \textbf{top-5} & \textbf{top-10} \\
    \midrule
    \multirow{3}{*}{ASR ($n{=}2048$)}
      & Base     & $32.10$ & $55.63$ & $64.24$ & $24.41$ & $47.50$ & $57.18$ \\
      & COAX     & $\mathbf{33.84}$ & $\mathbf{57.29}$ & $\mathbf{65.75}$ & $\mathbf{26.07}$ & $\mathbf{49.38}$ & $\mathbf{58.98}$ \\
      & $\Delta$ & $+1.74$ & $+1.66$ & $+1.51$ & $+1.66$ & $+1.88$ & $+1.80$ \\
    \midrule
    \multirow{3}{*}{Captioning ($n{=}1024$)}
      & Base     & $57.33$ & $81.98$ & $87.63$ & $50.44$ & $78.52$ & $85.20$ \\
      & COAX     & $\mathbf{58.34}$ & $\mathbf{82.89}$ & $\mathbf{88.29}$ & $\mathbf{51.49}$ & $\mathbf{79.57}$ & $\mathbf{85.99}$ \\
      & $\Delta$ & $+1.01$ & $+0.91$ & $+0.66$ & $+1.05$ & $+1.05$ & $+0.79$ \\
    \bottomrule
  \end{tabular}
  \caption{\textbf{COAX improves task-aligned ASR and captioning predictions.} Teacher-forced top-$k$ accuracy (\%) on reference transcripts and captions for a 345M-A147M model, evaluated without inference-time bias; higher is better. A word is counted as correct at $k$ only when all of its constituent tokens appear within the top-$k$ predictions. COAX improves all twelve metrics.}
  \label{tab:downstream-mm}
\end{table}

\paragraph{Free-running generation remains undertrained.} Neither model is sufficiently trained for meaningful free-running generation at this scale: greedy ASR exceeds $100\%$ WER for both models, while captioning CIDEr-D is near zero ($\approx0.032$ for both baseline and COAX), which indicates that neither model has yet learned reliable transcription or image grounding, making differences between the two arms difficult to interpret. We therefore use held-out validation loss as our primary metric and report teacher-forced task-level accuracy as supporting evidence where the models operate in a measurable regime.

\subsection{COAX in Continual Pretraining}
\label{app:cpt}

All main experiments apply COAX while pretraining from scratch. We additionally test whether it can allocate capacity to a new category introduced during continual pretraining. Starting from an unbiased 3B-A654M checkpoint at step $8000$ from Section~\ref{sec:results-multilingual}, we upcycle the model from $128$ to $160$ routed experts using the method described in \citet{dwivedi2026expert}. We then continue training with the new category comprising $20\%$ of tokens and biased toward the $32$ newly introduced experts.

Table~\ref{tab:cpt} shows that COAX improves continual pretraining on vision and code while also improving replay performance on the original four languages. The effect is less consistent for newly introduced languages, with Russian and Hindi showing essentially no improvement on the new-category loss. These results suggest that COAX can successfully direct newly introduced capacity toward new categories during continual pretraining, although its benefits depend on the category being introduced.


\begin{table}[b]
  \centering\small
  \setlength{\tabcolsep}{9pt}
  \begin{tabular}{lcc}
    \toprule
    \textbf{New category} & \textbf{New-category $\Delta$ loss (\%)} & \textbf{Replay $\Delta$ loss (\%)} \\
    \midrule
    Vision  & $\mathbf{-0.066}$ & $\mathbf{-0.067}$ \\
    Code    & $\mathbf{-0.177}$ & $\mathbf{-0.064}$ \\
    Russian & $+0.028$ & $\mathbf{-0.054}$ \\
    Hindi   & $+0.029$ & $\mathbf{-0.039}$ \\
    \bottomrule
  \end{tabular}
  \caption{\textbf{COAX can allocate newly added expert capacity during continual pretraining.} Relative change in validation loss ($\Delta\%$) against unbiased continual pretraining from the same checkpoint for the 3B-A654M model (from Section \ref{sec:results-multilingual}) upcycled from $128$ to $160$ routed experts at $\alpha=0.03$; negative is better. The new category receives $32/160$ experts and $20\%$ of training tokens. \emph{Replay} is the mean over the four original languages. }
  \label{tab:cpt}
\end{table}

\section{Analysis of COAX-Induced Specialization}
We next examine what COAX changes internally, first in the learned router distributions and then in the functional dependence of categories on their designated experts.

\subsection{Router Entropy under COAX}
\label{app:router-entropy}

\begin{figure}[t]
    \centering
    \includegraphics[width=0.6\textwidth]{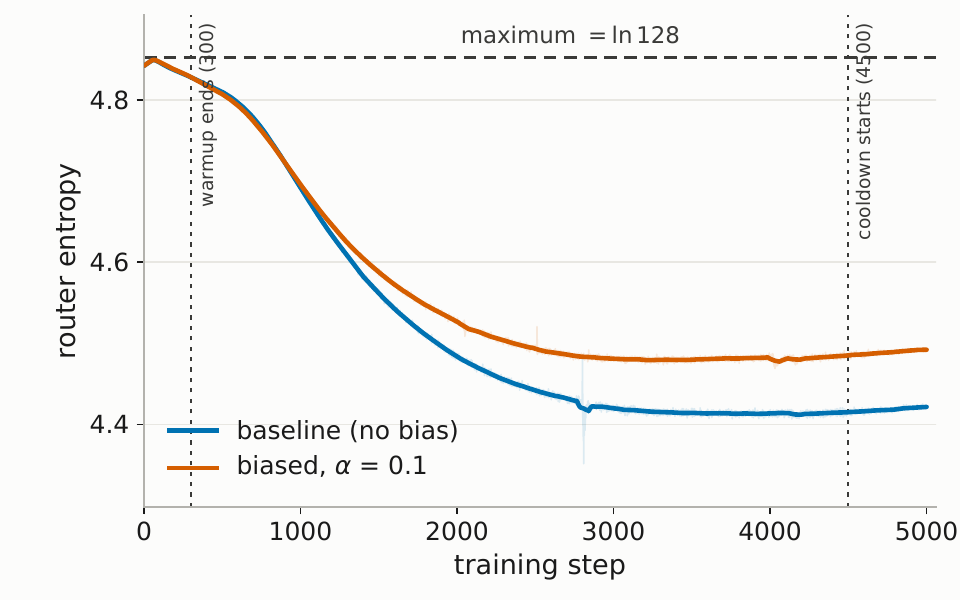}
    \caption{\textbf{COAX does not sharpen the learned router.} Router entropy over all $128$ experts, averaged across tokens and layers, for the 3B-A654M model on the four-language mixture from Section~\ref{sec:results-multilingual}. Entropy is computed before adding the category bias and therefore reflects the router's learned scores. Both models sharpen during training, but the COAX-trained router remains consistently flatter, suggesting that its performance gains arise from changes in expert learning rather than a better, more discriminate router.}
    \label{fig:router-entropy}
\end{figure}

Figure~\ref{fig:router-entropy} examines whether COAX improves performance simply by making the learned router more discriminative. Both models begin close to uniform routing entropy ($4.840$ against a maximum of $\ln 128=4.852$) and sharpen throughout training. However, the COAX-trained router sharpens \emph{less}, ending at $4.492$ compared with $4.422$ for the unbiased baseline and retaining only $\approx83\%$ of the baseline's entropy reduction. Since entropy is measured before adding the COAX bias, this reflects the router's learned distribution itself. The gains in Tables ~\ref{tab:ml-results} and ~\ref{tab:mm-results} therefore do not arise from a sharper router and instead suggest that COAX changes how the experts themselves are trained, and less categorical interference leads to more performant experts. We test this interpretation more directly in the following section by measuring whether category-specific functionality becomes localized in the designated experts.

\subsection{COAX Induces Functionally Modular Experts}
\label{app:modularity}

We test whether COAX induces functional modularity by removing the experts associated with a category and measuring what breaks. At every MoE layer, we mask four experts before top-$K$ selection and re-evaluate all four languages without inference-time bias. For COAX, we remove the four experts assigned to Swahili; for the unbiased baseline, which has no designated subset, we remove the four experts most frequently selected by Swahili tokens on a held-out slice of the training data. We analyze the same 3B-A654M models used in Section~\ref{sec:strength_alpha}.

Table~\ref{tab:modularity} shows that COAX concentrates Swahili-specific functionality into its designated experts. Removing four experts increases Swahili loss by only $9.34\%$ in the baseline, compared with $107.09\%$ for COAX at $\alpha=0.03$ and $128.18\%$ for annealed $\alpha=0.3$. Functional dependence grows alongside routing concentration: these experts receive $4.58\%$, $12.15\%$, and $26.96\%$ of Swahili's selections, respectively, compared with a $3.12\% (4/128)$ uniform-random rate. At annealed $\alpha=0.3$, the effect is also highly category-specific: removing the Swahili experts changes English, Korean, and Urdu loss by only $+0.08\%$, $+0.19\%$, and $+0.03\%$, respectively. In contrast, disabling arbitrary expert sets has little effect on Swahili, but more on English, Korean, and Urdu, suggesting greater specialization under COAX which is directly controllable by the strength of $\alpha$.


\begin{table}[t]
  \centering\small
  \setlength{\tabcolsep}{5pt}
  \begin{tabular}{llrcccc}
    \toprule
    \textbf{Disabled set} & \textbf{Model} & \textbf{Swahili traffic (\%)} & \textbf{Swahili} & \textbf{English} & \textbf{Korean} & \textbf{Urdu} \\
    \midrule
    \multirow{3}{*}{Swahili-targeted}
      & Baseline                  & $4.58$  & $+9.34$ & $+0.92$ & $+0.71$ & $+0.97$ \\
      & COAX $\alpha{=}0.03$      & $12.15$ & $+107.09$ & $+0.75$ & $+0.83$ & $+1.59$ \\
      & COAX $\alpha{=}0.3$ annealed  & $\mathbf{26.96}$ & $\mathbf{+128.18}$ & $\mathbf{+0.08}$ & $\mathbf{+0.19}$ & $\mathbf{+0.03}$ \\
    \midrule
    \multirow{3}{*}{Arbitrary}
      & Baseline                  & $2.99$ & $+0.87$ & $+0.93$ & $+0.76$ & $+0.68$ \\
      & COAX $\alpha{=}0.03$      & $2.70$ & $+0.54$ & $+0.87$ & $+0.86$ & $+0.47$ \\
      & COAX $\alpha{=}0.3$ annealed  & $2.10$ & $+0.25$ & $+0.78$ & $+0.90$ & $+0.04$ \\
    \bottomrule
  \end{tabular}
  \caption{\textbf{COAX concentrates Swahili-specific functionality into its designated experts.} Relative change in validation loss ($\Delta\%$) after disabling four of $128$ experts at every layer of the 3B-A654M model; larger positive values indicate greater functional dependence. \emph{Swahili traffic} is the fraction of Swahili top-$K$ selections routed through the disabled experts in the intact model; the uniform-random rate is $3.12\%$ ($4/128$). All models are evaluated without inference-time bias.}
  \label{tab:modularity}
\end{table}

\section{Comparisons with Alternative Specialization Methods}
We compare COAX with two alternative ways of inducing category-aligned specialization: enforcing category-specific expert partitions through hard routing \citep{gururangan-etal-2022-demix}, and explicitly separating category-level router distributions through an auxiliary objective \citep{ICLR2026_0c8f0ede}.

\subsection{DEMix-style Hard Routing}
\label{app:hard-routing}

DEMix hard-routes each domain through its dedicated expert during training \citep{gururangan-etal-2022-demix}. We generalize this baseline to our top-$K$ MoE by restricting each category to its designated expert subset, providing a direct hard-specialization counterpart to COAX. As expected, this yields an in-group routing fraction of exactly $1.0$. Table \ref{tab:hard-routing} shows the results. Removing the hard constraint at evaluation produces a catastrophic $+251.1\%$ regression on validation loss for the non-English languages, reflecting severe train-test mismatch rather than intrinsically poor learned weights. When evaluated under the same hard routing used during training, the minority mean is instead $+0.047\%$, approximately neutral relative to the unbiased baseline, with only three of eight languages improving.

Hard routing therefore provides no performance advantage despite substantially greater throughput cost. Its iteration-time overhead is $17.57\%$, compared with $1.79\%$ for soft COAX in the same 10.6B setup. COAX also attains a decrease of $0.286\%$ on validation loss for the minority languages while retaining cross-category routing. The comparison is not perfectly matched because the COAX column removes its bias at evaluation whereas the matched hard-routing column retains its constraint. However, retaining the COAX bias generally improves performance elsewhere in our experiments, so this asymmetry does not explain hard routing's lack of benefit. Overall, the result supports COAX's central design choice: partial specialization with sharing retained is preferable to enforcing complete category partitions.

\begin{table}[t]
  \centering\small
  \setlength{\tabcolsep}{6pt}
  \begin{tabular}{lrccc}
    \toprule
    \textbf{Language} & \textbf{Experts} & \textbf{COAX ($\alpha=0.03$)} & \textbf{Hard, weights-only} & \textbf{Hard + bias on eval} \\
    \midrule
    English  & $114$ & $-0.070$ & $+15.868$ & $+0.350$ \\
    Chinese  & $26$  & $-0.103$ & $+88.817$ & $+1.994$ \\
    Korean   & $26$  & $-0.104$ & $+118.728$ & $+1.679$ \\
    Khmer    & $30$  & $-0.003$ & $+604.163$ & $\mathbf{-4.090}$ \\
    Tamil    & $22$  & $-0.418$ & $+253.411$ & $\mathbf{-0.423}$ \\
    Georgian & $16$  & $-0.576$ & $+358.344$ & $+0.228$ \\
    Urdu     & $14$  & $-0.308$ & $+198.754$ & $\mathbf{-0.101}$ \\
    Swahili  & $8$   & $-0.486$ & $+135.243$ & $+1.039$ \\
    \midrule
    \multicolumn{2}{l}{\textbf{Mean, seven minorities}} & $\mathbf{-0.286}$ & $+251.066$ & $+0.047$ \\
    \midrule
    \multicolumn{2}{l}{\textbf{Iteration time}} & $+1.79\,\%$ & \multicolumn{2}{c}{$+17.57\,\%$} \\
    \bottomrule
  \end{tabular}
  \caption{\textbf{COAX outperforms generalized DEMix-style hard routing.} Relative change in validation loss ($\Delta\%$) against the same-seed unbiased baseline for the 10.6B-A1.2B eight-language setup from Section \ref{sec:8_language_result}; negative is better. \emph{Hard, weights-only} removes the routing bias at evaluation, whereas \emph{hard + bias on eval} retains it. Iteration-time overhead is measured using the same throughput probe as Appendix ~\ref{app:throughput}.}
  \label{tab:hard-routing}
\end{table}

\subsection{Expert Divergence Learning}
\label{app:edl}

Expert Divergence Learning (EDL) explicitly encourages category-aligned specialization by maximizing the pairwise Jensen-Shannon divergence between domain-level router distributions \citep{ICLR2026_0c8f0ede}. Unlike hard category routing, EDL leaves the routing architecture and inference procedure unchanged, making it a particularly relevant comparison to COAX. We test whether EDL's reported benefits on multi-domain mixtures transfer to our long-tailed multilingual mixture, where the central challenge is not only separating categories but allocating preferred expert capacity among categories with highly imbalanced data prevalence. We evaluate the coefficient used in the EDL paper, $\beta=5\times10^{-4}$, together with two smaller coefficients, $\beta=10^{-4}$ and $\beta=2\times10^{-5}$. All runs use the same training seed.

\paragraph{Increasing routing divergence alone does not improve the long tail.}
Table~\ref{tab:edl} compares EDL and COAX on the 10.6B-A1.2B eight-language mixture from Section~\ref{sec:8_language_result}. At $\beta=5\times10^{-4}$, EDL increases validation loss for every language, with a mean regression of $+1.016\%$ over the seven minority languages, whereas COAX improves the same mean by $-0.286\%$. The regression under EDL also tends to increase as language prevalence decreases: the Spearman correlation between data share and $\Delta$ loss is $\rho=-0.88$ across all eight languages ($-0.96$ excluding Khmer), with Swahili degrading by $+3.020\%$. Reducing the coefficient to $\beta=10^{-4}$ improves Korean, but the minority mean remains above the unbiased baseline. At $\beta=2\times10^{-5}$, the regression decreases to $+0.369\%$, but EDL still increases loss for every language. We therefore do not interpret category-level routing divergence itself as sufficient for improving low-resource performance in this setting.

\paragraph{EDL and COAX induce qualitatively different routing structure.}
Table~\ref{tab:edl-routing} measures the routing statistic optimized by EDL on the final checkpoints. The unbiased model has very low pairwise language-level router divergence ($0.0025$), indicating that its probability distributions are nearly indistinguishable across languages under this statistic. EDL changes this substantially: mean pairwise JSD increases to $0.0330$, $0.1560$, and $0.3299$ at $\beta=2\times10^{-5}$, $10^{-4}$, and $5\times10^{-4}$, respectively. COAX, in contrast, leaves probability-level divergence nearly unchanged at $0.0027$. This difference follows naturally from the mechanisms: EDL directly optimizes the router distributions, whereas COAX adds its category bias only for top-$K$ selection and retains the unbiased scores for combining selected experts.

The stronger router divergence under EDL is accompanied by a larger load-balancing objective: the normalized auxiliary loss increases from $1.0345$ for the baseline to $1.0581$, $1.1063$, and $1.2335$ as the EDL coefficient increases, while COAX leaves it essentially unchanged at $1.0346$. We additionally report $\mathrm{MaxVio}_{\mathrm{batch}}$ following \citet{wang2024auxiliary}; COAX yields a higher $\mathrm{MaxVio}_{\mathrm{batch}}$ than EDL at all three coefficients despite barely changing the underlying probability-level JSD. Together, these results sharpen the distinction between the two approaches: EDL explicitly optimizes for category-level routing divergence, whereas COAX directly controls the expert subset toward which each category is preferentially encouraged. In our long-tailed setting, the latter improves low-resource performance even though the router distributions themselves remain nearly as homogeneous as in the unbiased model. This suggests that specialization alone is not sufficient; how preferred expert capacity is allocated across categories also matters.


\begin{table}[t]
  \centering\small
  \setlength{\tabcolsep}{5pt}
  \begin{tabular}{lrcccc}
    \toprule
    \textbf{Language} & \textbf{Share} & \textbf{COAX ($\alpha=0.03$)} & \textbf{EDL ($\beta=5\!\times\!10^{-4}$)} & \textbf{EDL ($\beta=10^{-4}$)} & \textbf{EDL ($\beta=2\!\times\!10^{-5}$)} \\
    \midrule
    English  & $63.2\%$ & $\mathbf{-0.070}$ & $+0.047$ & $+0.082$ & $+0.065$ \\
    Chinese  & $10.0\%$ & $\mathbf{-0.103}$ & $+0.222$ & $+0.028$ & $+0.098$ \\
    Korean   & $10.0\%$ & $\mathbf{-0.104}$ & $+0.091$ & $-0.013$ & $+0.063$ \\
    Khmer    & $6.0\%$  & $\mathbf{-0.003}$ & $+1.193$ & $+3.866$ & $+0.738$ \\
    Tamil    & $4.2\%$  & $\mathbf{-0.418}$ & $+0.400$ & $+0.319$ & $+0.336$ \\
    Georgian & $3.0\%$  & $\mathbf{-0.576}$ & $+1.257$ & $+0.523$ & $+0.502$ \\
    Urdu     & $2.7\%$  & $\mathbf{-0.308}$ & $+0.926$ & $+0.333$ & $+0.174$ \\
    Swahili  & $0.85\%$ & $\mathbf{-0.486}$ & $+3.020$ & $+2.174$ & $+0.669$ \\
    \midrule
    \multicolumn{2}{l}{\textbf{Mean, seven minorities}} & $\mathbf{-0.286}$ & $+1.016$ & $+1.033$ & $+0.369$ \\
    \multicolumn{2}{l}{\textbf{Mean, excl. Khmer}}      & $\mathbf{-0.333}$ & $+0.986$ & $+0.561$ & $+0.307$ \\
    \bottomrule
  \end{tabular}
  \caption{\textbf{EDL does not improve the long-tailed multilingual mixture at any of the tested coefficients.} Relative change in validation loss ($\Delta\%$) against the matched unbiased baseline for the 10.6B-A1.2B eight-language setup of Section~\ref{sec:8_language_result}; negative is better. \emph{Share} denotes each language's fraction of the training mixture.}
  \label{tab:edl}
\end{table}

\begin{table}[t]
  \centering\small
  \setlength{\tabcolsep}{6pt}
  \begin{tabular}{lcccc}
    \toprule
    \textbf{Method} & \textbf{Aux.\ loss} & \textbf{Excess over uniform} & $\mathbf{\mathrm{MaxVio}_{\mathrm{batch}}}$ & \textbf{Router JSD} \\
    \midrule
    Unbiased baseline                 & $1.0345$ & $3.45\%$  & $0.4184$ & $0.0025$ \\
    COAX ($\alpha=0.03$)              & $1.0346$ & $3.46\%$  & $0.6568$ & $0.0027$ \\
    EDL ($\beta=2\!\times\!10^{-5}$)  & $1.0581$ & $5.81\%$  & $0.4618$ & $0.0296$ \\
    EDL ($\beta=10^{-4}$)             & $1.1063$ & $10.63\%$ & $0.5606$ & $0.1406$ \\
    EDL ($\beta=5\!\times\!10^{-4}$)  & $1.2335$ & $23.35\%$ & $0.4671$ & $0.3300$ \\
    \bottomrule
  \end{tabular}
  \caption{\textbf{EDL strongly separates language-level router distributions, while COAX leaves the probability-level divergence nearly unchanged.} Statistics are averaged over the final $100$ training steps, except \emph{Router JSD}, which is measured on the final checkpoint. The auxiliary loss is normalized such that uniform routing gives $1$. $\mathrm{MaxVio}_{\mathrm{batch}}$ \citep{wang2024auxiliary} measures the relative excess load of the most-loaded expert over its load under perfect balance. Router JSD is the mean pairwise Jensen-Shannon divergence between language-level router distributions, with maximum $\ln 2$.}
  \label{tab:edl-routing}
\end{table}

\section{Design Ablations and Hyperparameter Robustness}
We test whether COAX's gains depend sensitively on its main design choices or on the training regime, varying the preferred expert allocation, the layers receiving the bias, the strength of load balancing, and the training horizon.

\subsection{Sensitivity to the Relative Expert Share}
\label{app:rc-ablation}

Table~\ref{tab:mm-results-app} shows that allocating only proportional expert capacity to lower-resource categories ($r_c=1$) consistently underperforms $r_c=2$, i.e. $2\times$ preferred expert share for vision and audio. 

Furthermore, we test whether COAX is sensitive to the precise size of this share at $r_c\geq2$. We try this in a multilingual setting, varying only the Swahili expert subset from $2$ to $32$ experts while keeping the Korean and Urdu allocations fixed. As shown in Table~\ref{tab:rc-ladder}, held-out performance changes little across this wide range, with no consistent trend as the number of assigned experts increases. Thus, while allocating additional capacity is beneficial, COAX does not appear to require precise tuning of the expert-set size within the range tested.

\begin{table}[t]
  \centering\small
  \setlength{\tabcolsep}{5pt}
  \begin{tabular}{llcccccc}
    \toprule
    \textbf{Model} & \textbf{Method} & \textbf{T} & \textbf{I2T} & \textbf{T2I} & \textbf{AA} & \textbf{AT} & \textbf{ATA} \\
    \midrule
    \multirow{4}{*}{345M-A147M} & $r_c\!=\!1$ & $-0.28$ & $-0.30$ & $-0.22$ & $-0.23$ & $-0.20$ & $-0.35$ \\
     & + bias at eval & $-0.41$ & $-0.34$ & $-0.25$ & $-0.25$ & $-0.21$ & $-0.37$ \\
     & $r_c=2$ & $\mathbf{-0.78}^{*}$ & $\mathbf{-0.72}^{**}$ & $\mathbf{-0.56}^{**}$ & $\mathbf{-0.48}$ & $\mathbf{-0.38}$ & $\mathbf{-0.65}$ \\
     & + bias at eval & $\mathbf{-0.88}^{*}$ & $\mathbf{-0.76}^{**}$ & $\mathbf{-0.59}^{**}$ & $\mathbf{-0.50}^{*}$ & $\mathbf{-0.39}$ & $\mathbf{-0.67}$ \\
    \midrule
    \multirow{4}{*}{3B-A654M} & $r_c\!=\!1$ & $+0.47^{*}$ & $-0.17^{**}$ & $-0.14^{**}$ & $+0.13$ & $+0.16^{**}$ & $+0.18^{*}$ \\
     & + bias at eval & $+0.39^{*}$ & $-0.20^{**}$ & $-0.17^{**}$ & $+0.11$ & $+0.15^{**}$ & $+0.17^{*}$ \\
     & $r_c=2$ & $+0.23^{**}$ & $\mathbf{-0.25}^{**}$ & $\mathbf{-0.19}^{**}$ & $+0.04$ & $+0.11$ & $+0.09$ \\
     & + bias at eval & $+0.18^{**}$ & $\mathbf{-0.29}^{**}$ & $\mathbf{-0.22}^{**}$ & $+0.02$ & $+0.09$ & $+0.08$ \\
    \midrule
    \multirow{4}{*}{10.6B-A1.2B} & $r_c\!=\!1$ & $+0.02$ & $-0.15$ & $-0.11$ & $-0.07$ & $-0.09$ & $-0.07$ \\
     & + bias at eval & $-0.07$ & $-0.19$ & $-0.15$ & $-0.11^{*}$ & $-0.11^{*}$ & $-0.10$ \\
     & COAX ($r_c=2$) & $\mathbf{-0.05}$ & $\mathbf{-0.25}^{*}$ & $\mathbf{-0.18}^{*}$ & $\mathbf{-0.08}$ & $\mathbf{-0.09}$ & $\mathbf{-0.09}$ \\
     & + bias at eval & $\mathbf{-0.13}$ & $\mathbf{-0.31}^{**}$ & $\mathbf{-0.23}^{**}$ & $\mathbf{-0.12}^{*}$ & $\mathbf{-0.12}^{*}$ & $\mathbf{-0.13}$ \\
    \bottomrule
  \end{tabular}
  \caption{\textbf{COAX helps multimodal performance.} Mean relative change in validation loss ($\Delta\%$) against unbiased baselines trained with matched seeds; negative is better. Columns correspond to held-out text (T), image-to-text captioning (I2T), text-to-image generation (T2I), audio-only (AA), audio-to-text ASR (AT), and interleaved audio-text (ATA). For $r_c=1$, each modality is assigned an expert fraction matching its token share. For $r_c=2$, vision and audio are each biased toward approximately $2\times$ their token share, with text receiving the residual expert subset ($r_c\approx0.7$). \emph{'+~bias~at~eval'} retains the category-conditioned bias during inference; otherwise, it is removed. $^{*}$ and $^{**}$ denote paired differences whose $90\%$ and $95\%$ confidence intervals respectively exclude zero. Within each evaluation setting, \textbf{bold} identifies the better of $r_c\in\{1,2\}$ when COAX improves over the unbiased baseline ($\Delta < 0$).}
  \label{tab:mm-results-app}
\end{table}

\begin{table}[t]
  \centering\small
  \setlength{\tabcolsep}{7pt}
  \begin{tabular}{rcccc}
    \toprule
    \textbf{Swahili experts} & \textbf{Swahili} & \textbf{English} & \textbf{Korean} & \textbf{Urdu} \\
    \midrule
    $2$  & $+0.043 \pm 0.249$ & $-0.026 \pm 0.044$ & $-0.001 \pm 0.223$ & $+0.088 \pm 0.347$ \\
    $4$  & \multicolumn{4}{c}{\textit{reference}} \\
    $8$  & $-0.087 \pm 0.158$ & $-0.032 \pm 0.062$ & $-0.070 \pm 0.100$ & $-0.027 \pm 0.143$ \\
    $16$ & $+0.138 \pm 0.119$ & $-0.002 \pm 0.029$ & $-0.021 \pm 0.131$ & $+0.049 \pm 0.253$ \\
    $32$ & $+0.207 \pm 0.215$ & $+0.010 \pm 0.045$ & $+0.061 \pm 0.080$ & $+0.089 \pm 0.194$ \\
    \bottomrule
  \end{tabular}
  \caption{\textbf{COAX is robust to the exact relative expert share.} Varying only the number of experts assigned to Swahili while keeping Korean and Urdu fixed for the 3B-A654M four-language setup from Section \ref{sec:results-multilingual} at $\alpha=0.03$. Values report $\Delta\%$ relative to the $4$-expert configuration at the same seed, averaged over three seeds $\pm$ standard deviation; negative is better.}
  \label{tab:rc-ladder}
\end{table}

\subsection{Biasing All Layers Is a Robust Default}
\label{app:layers}

In our experiments, routing becomes increasingly language-selective with depth. Across all three model scales, the mean pairwise Jensen--Shannon divergence between language-specific expert distributions (measured on the validation sets) rises with layer depth (Spearman $\rho=0.69$--$0.97$), with minima around layers $3$--$9$ and maxima at layers $31$--$32$. This motivates two competing strategies: bias later layers where specialization already emerges, or bias less-selective earlier/middle layers where additional structure may be most useful. We test both by restricting COAX to contiguous layer ranges. Table~\ref{tab:layers} shows that none improves over applying COAX throughout the network. Biasing all layers gives the best four-language mean ($-0.158\%$), compared with $-0.029\%$ for late layers, $+0.105\%$ for early layers, and $+0.382\%$ for the middle block. 


\begin{table}[t]
  \centering\small
  \setlength{\tabcolsep}{5pt}
  \begin{tabular}{lccccc}
    \toprule
    \textbf{Biased layers} & \textbf{English} & \textbf{Korean} & \textbf{Urdu} & \textbf{Swahili} & \textbf{Mean} \\
    \midrule
    All ($1$--$32$)     & $+0.041$ & $+0.031 $ & $\mathbf{-0.142}$ & $\mathbf{-0.563} $ & $\mathbf{-0.158}$ \\
    Early ($1$--$21$)   & $+0.134 $ & $+0.215$ & $+0.164$ & $\mathbf{-0.093} $ & $+0.105 $ \\
    Middle ($12$--$21$) & $+0.346 $ & $+0.465 $ & $+0.405 $ & $+0.310 $ & $+0.382 $ \\
    Late ($22$--$32$)   & $+0.092 $ & $+0.069 $ & $+0.056 $ & $\mathbf{-0.333} $ & $\mathbf{-0.029} $ \\
    \bottomrule
  \end{tabular}
  \caption{\textbf{Restricting COAX to specific layer ranges does not improve over biasing all layers.} Relative change in validation loss ($\Delta\%$) against matched-seed unbiased baselines for the 345M-A147M four-language setup at $\alpha=0.03$, averaged over three seeds 28--30; negative is better. The mean is over all four languages.} 
  \label{tab:layers}
\end{table}

\subsection{Interaction with Load Balancing}
\label{app:aux-interaction}

COAX modifies expert selection while the auxiliary load-balancing loss encourages uniform expert utilization, raising the question of whether the two objectives interfere. All main experiments use an auxiliary coefficient of $10^{-3}$; here we vary it from $10^{-4}$ to $10^{-2}$ on the 3B-A654M four-language setup. Since changing this coefficient also affects the unbiased model, we train a matched baseline at each value and report COAX relative to the corresponding baseline. We test both the default $\alpha=0.03$ and the stronger $\alpha=0.3$ schedule annealed to zero during cooldown.

Table~\ref{tab:aux-interaction} shows that increasing the auxiliary coefficient to $1e^{-2}$ reduces the minority-language gain from $-0.42\%$ to $-0.18\%$ for $\alpha=0.03$, and from $-1.26\%$ to $-0.43\%$ for annealed $\alpha=0.3$. In contrast, the stronger annealed bias is essentially unchanged between coefficients $1e^{-4}$ and $1e^{-3}$, reaching $-1.26\%$ in both cases. The weaker $\alpha=0.03$ arm varies more across this range, probably because lower load-balancing regularization allows larger router logits, and $\alpha=0.03$ cannot affect the routing dynamics at an auxiliary coefficient of $1e^{-4}$. 

\begin{table}[t]
  \centering\small
  \setlength{\tabcolsep}{6pt}
  \begin{tabular}{llccccc}
    \toprule
    \textbf{Bias strength} & \textbf{Aux coeff.} & \textbf{English} & \textbf{Korean} & \textbf{Urdu} & \textbf{Swahili} & \textbf{Minority mean} \\
    \midrule
    \multirow{3}{*}{$\alpha=0.03$}
      & $1e^{-4}$ & $+0.17$ & $\mathbf{-0.07}$ & $\mathbf{-0.14}$ & $\mathbf{-0.33}$ & $\mathbf{-0.18}$ \\
      & $1e^{-3}$ & $\mathbf{-0.03}$ & $\mathbf{-0.26}$ & $\mathbf{-0.59}$ & $\mathbf{-0.41}$ & $\mathbf{-0.42}$ \\
      & $1e^{-2}$ & $+0.03$ & $\mathbf{-0.06}$ & $+0.02$ & $\mathbf{-0.51}$ & $\mathbf{-0.18}$ \\
    \midrule
    \multirow{3}{*}{$\alpha=0.3$ annealed}
      & $10^{-4}$ & $+0.48$ & $\mathbf{-0.40}$ & $\mathbf{-1.69}$ & $\mathbf{-1.68}$ & $\mathbf{-1.26}$ \\
      & $10^{-3}$ & $+0.57$ & $\mathbf{-0.49}$ & $\mathbf{-1.82}$ & $\mathbf{-1.48}$ & $\mathbf{-1.26}$ \\
      & $10^{-2}$ & $+0.67$ & $+0.04$ & $\mathbf{-0.41}$ & $\mathbf{-0.91}$ & $\mathbf{-0.43}$ \\
    \bottomrule
  \end{tabular}
  \caption{\textbf{COAX remains effective under standard load balancing, but strong load-balancing suppresses it.} Relative change in validation loss ($\Delta\%$) against an unbiased baseline trained with the same auxiliary coefficient for the 3B-A654M four-language setup from Section \ref{sec:results-multilingual}; negative is better. The minority mean averages Korean, Urdu, and Swahili. All evaluations remove the inference-time bias.}
  \label{tab:aux-interaction}
\end{table}

\paragraph{COAX barely changes the auxiliary load-balancing objective.} We additionally inspect the auxiliary load-balancing loss of the multimodal runs in Table~\ref{tab:mm-results}. Table~\ref{tab:aux-rc} shows that COAX changes this objective by at most $0.31\%$ across all model scales and allocations, despite the substantial category-aligned routing induced by COAX (Figure~\ref{fig:in-group-routing-fraction}). At 10.6B-A1.2B, for example, the auxiliary loss changes by only $-0.06\%$ at $r_c=1$ and $-0.22\%$ at $r_c=2$. Thus, COAX can impose meaningful category structure on expert selection while leaving the conventional load-balancing objective almost unchanged.

\begin{table}[b]
  \centering\small
  \setlength{\tabcolsep}{10pt}
  \begin{tabular}{lccc}
    \toprule
    \textbf{Model} & \textbf{Baseline} & \textbf{$r_c=1$} & \textbf{$r_c=2$} \\
    \midrule
    345M-A147M   & $1.0312$ & $+0.31\%$ & $+0.26\%$ \\
    3B-A654M     & $1.0202$ & $-0.08\%$ & $-0.24\%$ \\
    10.6B-A1.2B  & $1.0305$ & $-0.06\%$ & $-0.22\%$ \\
    \bottomrule
  \end{tabular}
  \caption{\textbf{COAX barely changes the auxiliary load-balancing loss.} Statistics are averaged over the final $100$ training steps of the multimodal runs in Table~\ref{tab:mm-results}. Baseline values are absolute auxiliary losses, while COAX columns report relative changes against the matched baseline, averaged over five seeds at 345M-A147M and four seeds at the larger scales.}
  \label{tab:aux-rc}
\end{table}

\subsection{COAX's Gains Persist with Longer Training}
\label{app:training-trajectory}

Our 10.6B-A1.2B results are only trained for 10B tokens, and the observed benefits of COAX might be an undertraining artifact, i.e., COAX may simply accelerate early optimization before an unbiased model eventually catches up. We test this directly using the two-group English-vs.-non-English configuration from Section~\ref{sec:coarse_grouping}, evaluating COAX and an unbiased baseline every $\approx1$B tokens (1000 steps) from $1$B to $10$B. Both runs use $\alpha=0.03$ and are evaluated without inference-time bias.

\begin{figure}[t]
  \centering
  \includegraphics[width=0.7\textwidth]{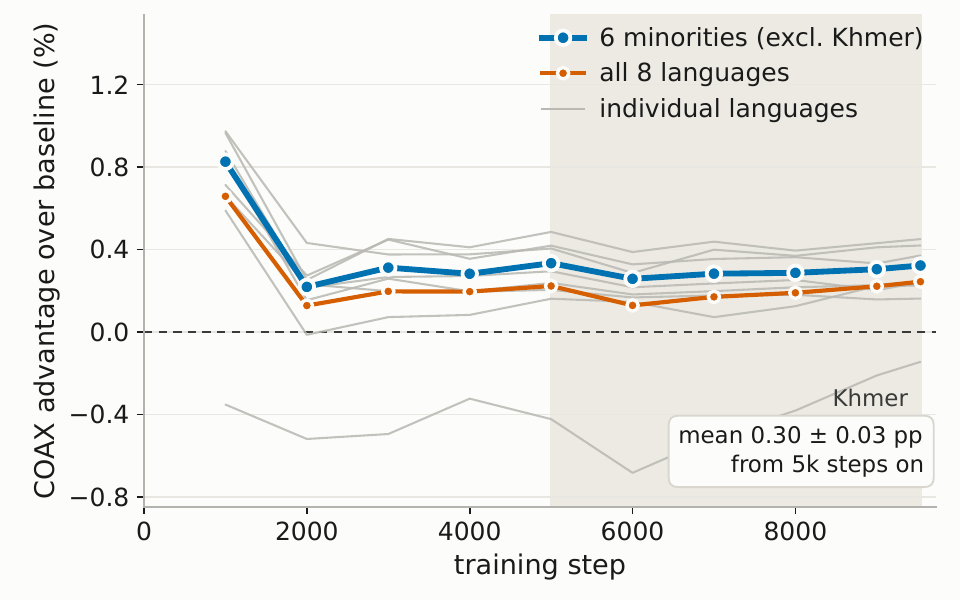}
  \caption{\textbf{COAX's gain does not decay over training steps.} Relative reduction in validation loss against the unbiased baseline at intermediate checkpoints of the 10.6B-A1.2B two-group run (Section~\ref{sec:coarse_grouping}); positive favors COAX. After an initially larger gain, the advantage stabilizes from approximately $5$B through the final $10$B-token checkpoint.}
  \label{fig:COAX-vs-training}
\end{figure}

Figure~\ref{fig:COAX-vs-training} shows that COAX's advantage is largest very early in training, decreases over the first few billion tokens, and then stabilizes rather than continuing to decay. From $5$B to $10$B tokens, the mean reduction in validation loss over the six non-Khmer minority languages is $0.298\pm0.028$ percentage points, with no downward trend. Khmer is the only consistent exception, showing a regression throughout training; its unusually low baseline loss under byte-fallback tokenization makes relative changes particularly unstable.

\begin{figure}[t]
  \centering
  \includegraphics[width=0.7\textwidth]{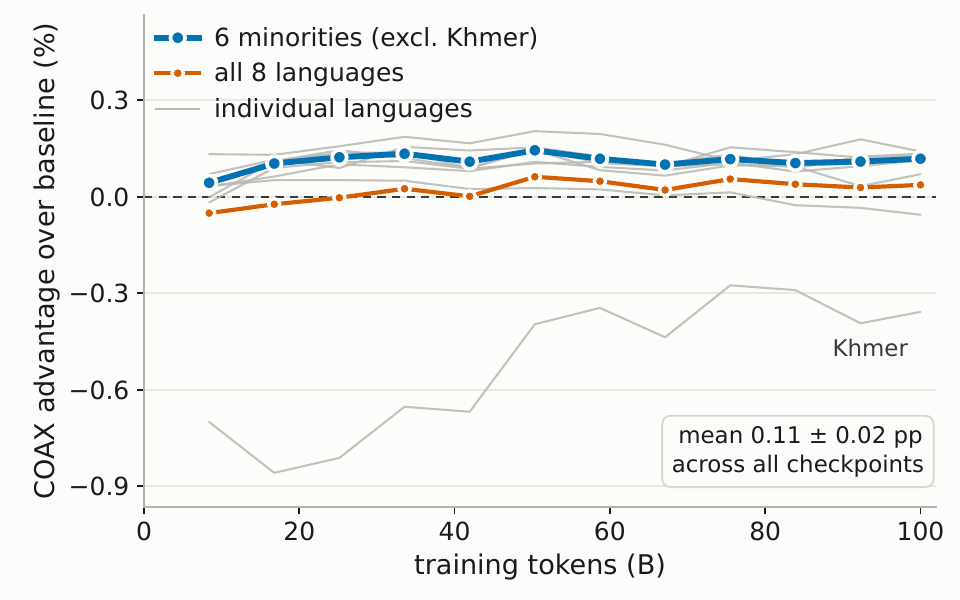}
  \caption{\textbf{COAX's advantage persists through 100B training tokens.} Relative reduction in validation loss against the unbiased baseline for the 10.6B-A1.2B two-group multilingual run; positive favors COAX. After increasing slightly from 10B to 20B tokens, the gain remains stable through the final 100B-token checkpoint.}
  \label{fig:COAX-vs-training-100B}
\end{figure}

We further extend this comparison to 100B training tokens (Figure~\ref{fig:COAX-vs-training-100B}) and find that COAX's gains remain stable over this substantially longer horizon. Reaching 100B tokens requires supplementing the training mixture with lower-quality multilingual data than used in our main experiments. The absolute gain at 10B tokens in this run is therefore not directly comparable to Figure~\ref{fig:COAX-vs-training}, and is smaller ($0.055$ vs.\ $0.312$ percentage points over the six non-Khmer minority languages). Within the 100B-token run, however, the gain increases slightly between 10B and 20B tokens and then remains stable, reaching $0.118$ percentage points at 100B. Together, the two training trajectories provide no indication that COAX's advantage decays away with continued training.

\section{Capacity Allocation versus Data Allocation}
COAX changes how expert capacity is preferentially allocated across categories without changing a given category's presence in the training data. We therefore ask whether its gains depend on separating the majority category, whether they can be reproduced by changing the data mixture itself, and whether COAX remains useful once conventional data upsampling is applied.

\subsection{Biasing the Majority Category Is Important}
\label{app:no-majority-bias}

COAX biases both minority and majority categories toward their assigned expert subsets. To test whether biasing the majority is necessary, we ablate the English bias in the same 10.6B-A1.2B eight-language configuration and expert allocation as Section~\ref{sec:8_language_result}, while retaining all seven minority-language biases. This also provides an extreme test of overlapping expert groups: removing the English bias is equivalent, for top-$K$ selection, to assigning English the entire expert pool, since adding the same bias to every expert leaves their relative routing scores unchanged. Table~\ref{tab:no-majority-bias} shows that this substantially weakens COAX: the ablation loses to full COAX on seven of eight languages and reduces the mean minority improvement from $-0.286\%$ to $-0.131\%$. English improves slightly more, but this small gain is accompanied by worse performance on every minority language. Thus, allowing the dominant category to freely overlap the minority expert subsets is less effective than maintaining distinct preferred capacity.

Together with the coarser grouping ablation in Table~\ref{tab:ml-practical-simple}, this suggests that reducing the number of expert groups is preferable to introducing overlap when fine-grained partitioning is not tractable because of too many categories. In particular, grouping all non-English languages into a shared minority subset performs well, whereas allowing English to overlap those minority subsets erodes much of the benefit. COAX therefore does not appear to require fine-grained category separation, but preserving separation between the dominant category and the long tail remains important.

The comparison uses a single seed, so we treat the magnitudes as directional rather than definitive. Khmer's large relative change is amplified by its unusually low baseline loss (as the tokenizer resorts to byte-fallback for Khmer), so its percentage change should not be compared directly with the absolute magnitudes of the other languages.

Having established that controlling the majority category's preferred capacity matters, we next ask whether the same effect can be obtained by reducing the majority category in the training data itself.

\begin{table}[t]
  \centering\small
  \setlength{\tabcolsep}{9pt}
  \begin{tabular}{lccc}
    \toprule
    \textbf{Language} & \textbf{COAX} & \textbf{No English bias} & \textbf{Difference} \\
    \midrule
    English  & $-0.070$ & $-0.090$ & $-0.020$ \\
    Chinese  & $-0.103$ & $-0.003$ & $+0.100$ \\
    Korean   & $-0.104$ & $+0.030$ & $+0.135$ \\
    Khmer    & $-0.003$ & $+0.729$ & $+0.733$ \\
    Tamil    & $-0.418$ & $-0.132$ & $+0.288$ \\
    Georgian & $-0.576$ & $-0.211$ & $+0.367$ \\
    Urdu     & $-0.308$ & $-0.071$ & $+0.238$ \\
    Swahili  & $-0.486$ & $-0.258$ & $+0.229$ \\
    \midrule
    \textbf{Mean, seven minorities} & $\mathbf{-0.286}$ & $-0.131$ & $+0.156$ \\
    \bottomrule
  \end{tabular}
  \caption{\textbf{Biasing the majority category is important.} Relative change in validation loss ($\Delta\%$) against the same-seed unbiased baseline for the 10.6B-A1.2B eight-language setup from Section~\ref{sec:8_language_result} at $\alpha=0.03$; negative is better. \emph{No English bias} retains the same expert allocation and minority biases as COAX but removes the routing bias for English. The final column is \emph{No English bias} minus \emph{COAX}, so positive values favor full COAX.}
  \label{tab:no-majority-bias}
\end{table}

\subsection{Reducing Majority Data Does Not Reproduce COAX}
\label{app:en-data-cut}

COAX gives English $114/256=44.53\%$ of the routed experts despite English comprising $63.24\%$ of the training mixture. One possible explanation is therefore that COAX merely down-weights the dominant category. We test this directly by reducing English to the same $44.53\%$ share in the \emph{data}, without applying any routing bias, while holding every minority language's absolute token count fixed. Table~\ref{tab:en-data-cut} shows the opposite behavior: COAX improves the minority validation loss by $-0.286\%$, whereas reducing English data degrades it by $+2.015\%$. Every language worsens in the data-reduction run. Thus, reducing majority data does not reproduce the effect of reducing the expert capacity towards which it is biased. 



This comparison has an unavoidable compute confound: holding minority token counts fixed while removing English reduces the total training tokens by $33.7\%$, so all shared parameters receive less training. We therefore use this experiment only to reject the simple equivalence between capacity reallocation and majority-data reduction, not to quantify their relative efficiency. 

A more practical alternative to reducing majority data is to increase the representation of the long tail, which we consider next.

\begin{table}[t]
  \centering\small
  \setlength{\tabcolsep}{7pt}
  \begin{tabular}{lcc}
    \toprule
    \textbf{Language} & \textbf{COAX: English gets $44.5\%$ of experts} & \textbf{English reduced to $44.5\%$ of data} \\
    \midrule
    English  & $\mathbf{-0.070}$ & $+6.583$ \\
    Chinese  & $\mathbf{-0.103}$ & $+2.306$ \\
    Korean   & $\mathbf{-0.104}$ & $+1.822$ \\
    Khmer    & $\mathbf{-0.003}$ & $+1.045$ \\
    Tamil    & $\mathbf{-0.418}$ & $+1.864$ \\
    Georgian & $\mathbf{-0.576}$ & $+2.554$ \\
    Urdu     & $\mathbf{-0.308}$ & $+1.925$ \\
    Swahili  & $\mathbf{-0.486}$ & $+2.589$ \\
    \midrule
    \textbf{Mean, seven minorities} & $\mathbf{-0.286}$ & $+2.015$ \\
    \bottomrule
  \end{tabular}
  \caption{\textbf{Reducing majority data and reducing majority expert capacity are not the same.} Relative change in validation loss ($\Delta\%$) against the same-seed unbiased baseline for the 10.6B-A1.2B eight-language setup from Section \ref{sec:8_language_result}; negative is better. In the data-reduction run, every minority retains its original absolute token count while only English is reduced. Reducing English data regresses performance across all languages, unlike COAX.}
  \label{tab:en-data-cut}
\end{table}

\subsection{COAX and Data Upsampling Are Complementary}
\label{app:additivity}

Data upsampling is a strong baseline for long-tailed mixtures, so we test whether COAX provides additional gains once the low-resource data itself has already been upsampled. In the 3B four-language setup (Section~\ref{sec:results-multilingual}), we double Urdu and Swahili to exactly $2$ epochs, increasing training compute by $3.61\%$. As shown in Table~\ref{tab:additivity}, data upsampling alone produces a substantially larger effect than COAX alone ($-3.579\%$ vs.\ $-0.260\%$ on the minority mean). Combining the two reaches $-3.738\%$, corresponding to an additional $-0.159\%$ improvement over data upsampling alone and approximately $61\%$ of COAX's standalone benefit.


\begin{table}[t]
  \centering\small
  \setlength{\tabcolsep}{6pt}
  \begin{tabular}{lccccc}
    \toprule
    \textbf{Arm} & \textbf{English} & \textbf{Korean} & \textbf{Urdu} & \textbf{Swahili} & \textbf{Minority mean} \\
    \midrule
    COAX only & $+0.011$ & $\mathbf{-0.063}$ & $\mathbf{-0.364}$ & $\mathbf{-0.352}$ & $\mathbf{-0.260}$ \\
    Data only ($2\times$ tail) & $+0.276$ & $+0.273$ & $\mathbf{-5.317}$ & $\mathbf{-5.694}$ & $\mathbf{-3.579}$ \\
    Both & $+0.274$ & $+0.126$ & $\mathbf{-5.420}$ & $\mathbf{-5.921}$ & $\mathbf{-3.738}$ \\
    \bottomrule
  \end{tabular}
  \caption{\textbf{COAX and data upsampling are additive.} Relative change in validation loss ($\Delta\%$) against matched-seed unbiased baselines for the 3B-A654M four-language setup at $\alpha=0.03$, averaged over three seeds; negative is better. The data run doubles Urdu and Swahili to $2$ epochs at $+3.61\%$ training compute. The minority mean averages Korean, Urdu, and Swahili.}
  \label{tab:additivity}
\end{table}

\section{Computational Efficiency}
We evaluate COAX's computational cost from two complementary perspectives: its measured training-time overhead and the amount of additional baseline training required to recover the same validation loss.

\subsection{Training Throughput Overhead of COAX}
\label{app:throughput}

The only systems optimization in our implementation is to distribute each biased expert subset evenly across expert-parallel groups. We measure the training overhead introduced by COAX after the routing dynamics have stabilized. For each model, we load the checkpoint at step 8000 and average TFLOP/s and time per iteration over steps 8050-8200. All measurements use the same 64 GPUs to ensure comparability. We use an A/B/B/A measurement protocol, where A denotes the unbiased baseline and B denotes COAX; this ordering mitigates monotonic drift during the allocation, while the repeated inner measurements provide a same-node repeatability estimate.

\begin{table}[t]
  \centering\small
  \setlength{\tabcolsep}{8pt}
  \begin{tabular}{lcccc}
    \toprule
    & \multicolumn{2}{c}{\textbf{Multimodality}} & \multicolumn{2}{c}{\textbf{Multilinguality}} \\
    \cmidrule(lr){2-3}\cmidrule(lr){4-5}
    \textbf{Model} & $\Delta$ TFLOP/s & $\Delta$ time/iter & $\Delta$ TFLOP/s & $\Delta$ time/iter \\
    \midrule
    345M-A147M   & $-17.0\%$ & $+20.2\%$ & $-17.7\%$ & $+21.9\%$ \\
    3B-A654M     & $-11.6\%$ & $+13.2\%$ & $-13.2\%$ & $+15.1\%$ \\
    10.6B-A1.2B  & $-2.4\%$  & $+2.4\%$  & $-1.7\%$  & $+1.7\%$ \\
    10.6B-A1.2B (Coarser)  & -  &  -  & $-1.3\%$  & $+1.3\%$ \\
    \bottomrule
  \end{tabular}
  \caption{
  \textbf{Training throughput overhead of COAX} at $\alpha=0.03$, measured over steps 8050--8200 from the step-8000 checkpoint. Negative $\Delta$ TFLOP/s and positive $\Delta$ time/iter indicate slower training relative to the unbiased baseline. The overhead decreases sharply with model scale, reaching only $1.7$--$2.4\%$ at 10.6B-A1.2B. 'Coarser' denotes the coarser configuration of COAX from Section \ref{sec:coarse_grouping}.
  }
  \label{tab:throughput}
\end{table}

Table~\ref{tab:throughput} shows that the computational overhead of COAX decreases sharply with model scale. At 345M-A147M, COAX increases iteration time by approximately $20$--$22\%$, falling to $13$--$15\%$ at 3B-A654M and only $1.7$--$2.4\%$ at 10.6B-A1.2B. We hypothesize that this scaling arises because COAX creates more uneven per-expert token counts within grouped GEMMs, reducing utilization; as expert GEMMs become larger with model scale, this imbalance and the associated routing/dispatch overheads are better amortized. Importantly, COAX continues to provide performance improvements at the largest scale (Tables~\ref{tab:ml-results}  and ~\ref{tab:mm-results}) despite this substantially smaller overhead. The overhead is even smaller when a coarser expert configuration is used (Section \ref{sec:coarse_grouping}) despite keeping the performance improvements comparable. Thus, at 10.6B-A1.2B, COAX provides consistent gains at a small additional training cost. For comparison, Appendix~\ref{app:hard-routing} shows that enforcing hard category routing raises iteration time by $17.57\%$ at the same scale, further motivating the soft-bias formulation.

\subsection{How Much Additional Training Is COAX Worth?}
\label{app:data-equivalence}

Small relative changes in validation loss can be difficult to interpret in practical terms. We therefore ask an equivalent question: how much longer must an unbiased baseline train to reach the loss achieved by COAX? For a fixed model, data mixture, and seed, we hold the COAX checkpoint fixed and continue training the corresponding baseline at the same learning rate, periodically evaluating both on the same held-out sets. The first point at which the baseline reaches COAX gives the equivalent additional training.

\paragraph{At 10.6B, COAX is worth more training than it costs.} We first measure this in the practical two-group 10.6B-A1.2B eight-language configuration, where English forms one group and the remaining seven languages share another. COAX uses the pre-cooldown 8000th step ($\approx 8.4$B tokens) checkpoint for comparison. Table~\ref{tab:data-equivalence-10b} shows that the baseline requires $2.7\%$ additional training tokens on average to match the weights-only COAX checkpoint, increasing to $3.8\%$ when Khmer is excluded because COAX does not improve it. Keeping the COAX bias active at evaluation raises these values to $3.8\%$ and $4.5\%$, respectively. In comparison, this simpler configuration of COAX increases iteration time by only $1.3\%$ compared to $1.8\%$ of the finer expert allocation (Appendix \ref{app:throughput}), making its training-time benefit larger than its measured overhead.

\begin{table}[t]
  \centering\small
  \setlength{\tabcolsep}{6pt}
  \begin{tabular}{lrrr}
    \toprule
    \textbf{Language} & \textbf{Data \%} & \textbf{COAX benefit} & \textbf{Equivalent extra training} \\
    \midrule
    English  & $63.24$ & $-0.170$ & $+1.7\%$ \\
    Chinese  & $10.00$ & $-0.191$ & $+2.9\%$ \\
    Korean   & $10.00$ & $-0.275$ & $+3.7\%$ \\
    Khmer    & $6.00$  & $+0.254$ & --- \\
    Tamil    & $4.19$  & $-0.494$ & $+6.3\%$ \\
    Georgian & $2.99$  & $-0.527$ & $+5.8\%$ \\
    Urdu     & $2.73$  & $-0.334$ & $+3.4\%$ \\
    Swahili  & $0.85$  & $-0.194$ & $+0.7\%$ \\
    \midrule
    \textbf{Mean, all 8} & & $\mathbf{-0.241}$ & $\mathbf{+2.7\%}$ \\
    \textbf{Mean, excl. Khmer} & & $\mathbf{-0.312}$ & $\mathbf{+3.8\%}$ \\
    \bottomrule
  \end{tabular}
  \caption{\textbf{At 10.6B, COAX is worth $2.7\%$ additional baseline training.} Results for the 10.6B-A1.2B two-group eight-language configuration at $\alpha=0.03$, evaluated without inference-time bias. \emph{COAX benefit} is the relative change in validation loss ($\Delta\%$) at step $8000$ (no cooldown, unlike the final checkpoint numbers from Section \ref{sec:results-multilingual}); negative favors COAX. \emph{Equivalent extra training} is the additional number of baseline training tokens required to reach the corresponding COAX loss, obtained by continuing the baseline at the same learning rate and interpolating between evaluated checkpoints. Khmer has no equivalence point because COAX is worse on it at the reference checkpoint.}
  \label{tab:data-equivalence-10b}
\end{table}

\paragraph{The equivalence is larger at the smallest scale.} We repeat the analysis for a 345M-A147M multimodal model from Section \ref{sec:results-multimodal}. We train COAX and its baseline for $10$B tokens at a constant learning rate and then extend the baseline to $12.1$B tokens. Averaged across all six validation sets, the baseline requires $18.5\%$ additional training tokens to reach the weights-only COAX loss. Retaining the COAX bias at evaluation increases the equivalence to $20.8\%$. Four of six validation sets cross within the measured ladder, while the two vision sets remain ahead at $12.1$B tokens and therefore require extrapolation individually.

\begin{table}[t]
  \centering\small
  \setlength{\tabcolsep}{6pt}
  \begin{tabular}{lcccccc}
    \toprule
     & \textbf{T} & \textbf{I2T} & \textbf{T2I} & \textbf{AA} & \textbf{AT} & \textbf{ATA} \\
    \midrule
    COAX benefit at $10$B & $-0.613$ & $-0.531$ & $-0.399$ & $-0.498$ & $-0.206$ & $-0.471$ \\
    Baseline gap at $12.1$B & $\mathbf{-0.041}$ & $+0.081$ & $+0.046$ & $\mathbf{-0.171}$ & $\mathbf{-0.239}$ & $\mathbf{-0.165}$ \\
    \midrule
    Equivalent extra training & $+19.6\%$ & $+23.9\%^{\dagger}$ & $+23.5\%^{\dagger}$ & $+18.2\%$ & $+8.3\%$ & $+16.3\%$ \\
    \bottomrule
  \end{tabular}
  \caption{\textbf{At 345M, matching COAX requires substantially more baseline training.} Results for the 345M-A147M multimodal model at seed 29 under a constant learning rate, evaluated without inference-time bias. The first row gives COAX's relative validation-loss improvement at $10$B tokens. The second gives the extended baseline's gap relative to the fixed COAX checkpoint after $12.1$B tokens; negative means the baseline has overtaken COAX. $^{\dagger}$ denotes equivalence points extrapolated from the log-linear fit because I2T and T2I had not crossed within the measured training ladder.}
  \label{tab:data-equivalence}
\end{table}

The training equivalence decreases from $18.5\%$ at 345M-A147M to $2.7\%$ at 10.6B-A1.2B because additional training reduces loss much faster in the larger-model experiment. Importantly, COAX's throughput overhead falls even more sharply ($20.2\%$ to $2.4\%$), particularly in the case of coarser expert grouping from Section \ref{sec:coarse_grouping} ($2.4\%$ to $1.3\%$). Thus, COAX provides a favorable training-cost trade-off at the 10.6B-A1.2B scale, particularly with coarser expert grouping.